\documentclass[letterpaper]{article}
\usepackage{aaai20}
\usepackage{times}
\usepackage{helvet}
\usepackage{courier}
\usepackage{url}            
\usepackage{acronym}
\usepackage{multirow}
\usepackage{amsmath,amssymb,amsthm,amsfonts,bm,bbm}
\usepackage{algorithmic}
\usepackage{graphicx}
\usepackage{textcomp}
\usepackage{xcolor}
\usepackage{gensymb}
\usepackage[font=footnotesize]{subcaption}
\usepackage{lipsum}  
\usepackage{etoolbox}
\usepackage{comment}

\newtoggle{anonymous}
\settoggle{anonymous}{true} 

\newtoggle{comments}
\settoggle{comments}{false} 
\iftoggle{comments}{
  \newcommand {\alberto}[1]{{\color{orange}{~Alberto: #1}\normalfont}}
  \newcommand {\bjin}[1]{{\color{blue}{~Baihong: #1}\normalfont}}
  \newcommand {\yuxin}[1]{{\color{violet}{~Yuxin: #1}\normalfont}}
  \newcommand {\yingshui}[1]{{\color{cyan}{~Yingshui: #1}\normalfont}}
  \newcommand {\alex}[1]{{\color{cyan}{~Alex: #1}\normalfont}}
  \newcommand {\red}[1]{{\color{red}{#1}\normalfont}}
}{
  \newcommand {\alberto}[1]{{}}
  \newcommand {\bjin}[1]{{}}
  \newcommand {\yuxin}[1]{{}}
  \newcommand {\alex}[1]{{}}
  \newcommand {\yingshui}[1]{{}}
  \newcommand {\red}[1]{{}}
}

\newcommand{\expctover}[2]{\mathbb{E}_{#1}\!\left[#2\right]}
\newcommand{\entropy}[1]{\mathbb{H}\left[#1\right]}

\newcommand{\Alg}{\mathcal{A}}

\newcommand{\given}{\mid}

\acrodef{SVM}{Support Vector Machine}
\acrodef{AE}{AutoEncoder}
\acrodef{RNN}{Recurrent Neural Network}
\acrodef{CNN}{Convolutional Neural Network}
\acrodef{OC-SVM}{One-class Support Vector Machine}
\acrodef{CPS}{Cyber-Physical System}
\acrodef{FDD}{Fault Detection and Diagnosis}
\acrodef{LDA}{Linear Discriminant Analysis}
\acrodef{PCA}{Principal Component Analysis}
\acrodef{MSE}{Mean Square Error}
\acrodef{LAMPS}{Laser Additive Manufacturing Pilot System}
\acrodef{AM}{Additive Manufacturing}
\acrodef{SLS}{Selective Laser Sintering}
\acrodef{IR}{infrared}
\acrodef{IoT}{Internet of Things}
\acrodef{ROC}{Receiver Operating Characteristic}
\acrodef{RMS}{Root-Mean-Square}
\acrodef{AUC}{Area Under the Curve}
\acrodef{VAE}{Variational Autoencoder}
\acrodef{GAN}{Generative Adversarial Networks}
\acrodef{FC}{Fully Connected}

\begin{document}
%
\title{Augmenting Monte Carlo Dropout Classification Models with Unsupervised Learning Tasks for Detecting and Diagnosing Out-of-Distribution Faults}
\author{Baihong~Jin$^{\dagger1}$~~Yingshui~Tan$^{\dagger1,2}$~~{Yuxin Chen}$^{*3}$~~{Alberto~Sangiovanni-Vincentelli}$^{1}$\\
$^{1}$Department of EECS, University of California, Berkeley, USA\\
$^{2}$RWTH Aachen University, Germany\\
$^{3}$California Institute of Technology, USA
}
\maketitle
\begin{abstract}
\begin{quote}
  The Monte Carlo dropout method has proved to be a scalable and easy-to-use approach for estimating the uncertainty of deep neural network predictions. This approach was recently applied to \ac{FDD} applications to improve the classification performance on incipient faults. In this paper, we propose a novel approach of augmenting the classification model with an additional unsupervised learning task. We justify our choice of algorithm design via an information-theoretical analysis. 
  Our experimental results on three datasets from diverse application domains show that the proposed method leads to improved fault detection and diagnosis performance, especially on out-of-distribution examples including both incipient and unknown faults.
\end{quote}
\end{abstract}

\section{Introduction}\label{sec:intro}

\noindent Data-driven approaches relying on supervised deep learning have achieved considerable success in many application domains due to their ability to classify data from multiple classes. Although supervised deep learning methods tend to perform well on known (in-distribution) data patterns, the unseen (out-of-distribution) data may lead to unexpected prediction behaviors. In the context of \acf{FDD}, labeled data for normal (fault-free) states and high-severity fault states are more easily accessible. On the contrary, labeled data for incipient faults (low-severity faults of known types) are more difficult to obtain~\cite{jin2019one} and usually missing or underrepresented in the dataset. In addition, in real-world operation, there could be unknown faults that do not belong to any fault types modeled in a classification model. These out-of-distribution faults not seen by the model in the training phase may fool a classification model into wrong belief, which is undesirable for \ac{FDD} applications. Although this problem conceptually can be alleviated by training the model on a larger and more comprehensive dataset, in practice it is technically impossible to include data of all different fault types, and of all possible severity levels. It is desirable for a diagnostic algorithm to report---in addition to the decisions---uncertainty estimates behind the decisions. Highly uncertain cases could then be flagged as requiring particular attention in future maintenance and repair actions.

\looseness -1 For estimating a neural network's prediction uncertainty, the standard way is to use a Bayesian approach whose goal is to learn a distribution over the network weights; however, such approach is computationally expensive and has not been widely adopted yet. A recent seminal work~\cite{gal2016uncertainty} discovered that one can approximate the posterior distribution of a dropout neural network by repeatedly sampling its predictions with dropout turned on at test time. This method, referred to as Monte Carlo dropout (MC dropout), provides an efficient and scalable way to perform Bayesian inference that can easily fit into the standard training pipelines of today's deep learning frameworks. The MC dropout approach due to its ease of use has been applied to disease diagnosis~\cite{leibig2017leveraging} and time series anomaly detection~\cite{zhu2017deep}. \citeauthor{jin2019detecting}~\shortcite{jin2019detecting} extended the use of MC dropout into a multiclass setting, and showed that MC dropout can not only help detect incipient faults but is also able to give informative hints about the types of these difficult-to-diagnose faults with the produced uncertainty estimates.


Apart from the supervised classification approaches, one category of unsupervised methods called one-class models~\cite{tax2002one} are particularly appealing in anomaly detection and \ac{FDD} applications, because only normal (fault-free) data are required to train a detection model. A one-class model aims to learn the distribution of the data for the normal operating condition of a system. Outliers to the learned distribution are recognized as anomalies or faults.
Typical examples of one-class models include autoencoders~\cite{thompson2002implicit} and one-class support vector machines~\cite{scholkopf2001estimating}. In practice, it can sometimes be difficult to train a good one-class classification model~\cite{jin2019one}, due to the lack of fault data for cross-validation. On the other hand, even when fault data are available, there is no straightforward way to directly incorporate them into the training process. In addition, as these models can only tell whether or not an input data point belong to the normal data distribution, they lack the diagnostic ability to differentiate between faults of different types.

The aforementioned reasons motivate us to devise a method that can leverage the strengths of both supervised and unsupervised learning approaches. The resulting model should not be only good at classifying in-distribution data, but should also be able to give reasonable uncertainty estimates in addition to the diagnostic decisions when predicting out-of-distribution fault examples. We summarize our main contributions in this paper as follows.
\begin{itemize}
\item We propose a novel neural network architecture that combines an MC dropout classifier and an autoencoder, thereby leveraging the strengths of supervised and unsupervised learning into one joint-training framework.
\item \looseness -1 We motivate the design choice of regularizing the latent space representation with a ``decoding pathway'' from an information theory point of view. The experimental results match the conjecture from our theoretical analysis.
\item Our experimental results on three datasets from different domains have demonstrated superior \ac{FDD} performance compared to non-augmented MC dropout classifiers and MC-dropout autoencoders, especially on out-of-distribution fault examples.
\end{itemize}



\section{Problem Statement}

\paragraph{Fault Detection}
Let $\mathcal{X}$ be the set of data points, and $\mathcal{M}$ be a model class. Each $M\in\mathcal{M}$ defines an \textit{anomaly score} function $s^M:\mathcal{X}\rightarrow\mathbb{R}$ that characterizes how likely a data point corresponds to a fault state; a larger $s^M(x)$ implies higher chance of a data point $x$ being a fault. For a given threshold value $\tilde{s}>0$, we can define the \textit{precision} and \textit{recall} of the model $M$ on the test data distribution as follows:
\begin{align}
  &\text{precision}(s^M,\tilde{s}) = \expctover{}
    {\mathbbm{1}\{x\text{~is a fault}\} \mid {s^M(x)>\tilde{s}}},\notag\\
  &\text{recall}(s^M,\tilde{s}) = \expctover{}{\mathbbm{1}\{s^M(x)>\tilde{s}\} \mid x \text{~is a fault}}.\notag
\end{align}
Our goal is to learn a score function $s^\ast$ and a corresponding threshold $\tilde{s}$, such that $(s^\ast,\tilde{s})$ can optimize the precision and the recall on unseen test data.
\paragraph{Fault Diagnosis} The fault diagnosis problem can be viewed as a natural extension of the fault detection problem. A fault diagnosis model not only needs to detect the existence of faults, but also differentiate between faults of different classes. In this paper, we view a fault diagnosis model that can deal with a situation of $n$ fault classes as the aggregation of $n$ fault detection models, where each fault detection model aims to distinguish the normal class and one particular type of fault. Concretely, the anomaly score function $s^M$ outputs one anomaly score for each type of fault, and we aim at identifying the optimal score function $s^\ast: \mathcal{X}\rightarrow \mathbb{R}^n$ and threshold $\tilde{s}\in \mathbb{R}^n$ which result in the best precision and recall for all the faults. It is possible to assign multiple fault labels to a given input, which means the fault diagnosis problem in our context is not only a multiclass classification problem, but also a multilabel classification problem.

\section{Motivation of Algorithm Design}\label{sec:motivation}

We propose to use the prediction uncertainty on the label of the input data as our score function $M$ for identifying fault data points. In the following, we motivate our choice of score function from an information-theoretic perspective.

Let us assume a classification model $M_\mathcal{A}$ that is trained with learning algorithm $\mathcal{A}$ on training set $\mathcal{D}_\text{train}$. The training set consists of data of class ``normal'' (labeled 0) and of class ``fault'' (labeled 1). In the test set $\mathcal{D}_\text{test}$, besides data that are in the training distribution, there are also data that do not belong to the training distribution. We denote the index set of normal data, fault data, and out-of-distribution data in the test set respectively as $\mathcal{I}_0$, $\mathcal{I}_1$, and $\mathcal{I}_2$.

Given input $x_i$, the output of the trained model $M_\mathcal{A}$ is a random variable $Y_i$. 
Let $\entropy{Y} = -\sum_y p(y)\log p(y)$ be the entropy of random variable $Y$ under distribution $p$. By the conditional independence of data points on model, we can decompose the total entropy of the output variable on the test set into three parts,
\begin{align}
  &\entropy{\mathcal{D}_\text{test} \given M_\mathcal{A}} = \underbrace{\sum_{i\in\mathcal{I}_0}\entropy{Y_i\,\vert\,M_\mathcal{A}}}_{P_0:~\text{entropy of normal data}} \notag\\
  &\quad\quad + \underbrace{\sum_{i\in\mathcal{I}_1}\entropy{Y_i\,\vert\,M_\mathcal{A}} + \sum_{i\in\mathcal{I}_2}\entropy{Y_i\,\vert\,M_\mathcal{A}}.}_{P_1:~\text{entropy of in-distribution and out-of-distribution fault data}}\label{eq:entropy_decomp}
\end{align}

Since we want to utilize prediction uncertainty as an indicator for identifying fault data points (especially the out-of-distribution faults), it is our desire to devise a learning algorithm to drive down the entropy (uncertainty) on normal data points, and meanwhile to increase the uncertainty on the out-of-distribution faults so that they are more distinguished from the in-distribution data points.
To lower the uncertainty on a selected subset of the data points (in our case the normal data points), a basic idea is to devote more effort to this class during training. One straightforward way is to increase the weight for the data points in the normal class when training the classification model; however, this may not yield the desired result because a large class weight will push the decision boundary farther away from the normal class and thus allowing more \emph{incipient fault} data points to be mistakenly classified as normal with little uncertainty.

To bypass this problem and better detect incipient faults, we propose to use an auxiliary task at a hidden layer for regularization purpose, which will encourage the network to learn different latent space representations for the data at this hidden layer. Since part of the network's capacity is devoted to the auxiliary task, it can be expected that the entropy at the output of the resulting classifier on the test data is increased with the modified training algorithm $\mathcal{A}'$, i.e.
\begin{align}
  \entropy{\mathcal{D}_\text{test}\given M_\mathcal{A}} \leq \entropy{\mathcal{D}_\text{test} \given M_\mathcal{A}'}.\notag
\end{align}
Here apostrophes are used to indicate variables that correspond to the modified training algorithm $\mathcal{A}'$. By Eq.~\eqref{eq:entropy_decomp}, $   \entropy{\mathcal{D}_\text{test}\given M_\mathcal{A}} = P_0 + P_1$; hence we have 
\begin{align}
  P_0' \leq P_0 \Rightarrow P_1 \leq P_1'. \notag
\end{align}
The above relation suggests that, for algorithm $\Alg'$, a lower uncertainty on the normal data points comes at the price of higher uncertainty on the in-distribution fault data points.
This analysis, however, does not give a definitive conclusion on the entropy of out-of-distribution (incipient and unknown) faults. We conjecture that these out-of-distribution examples will also exhibit higher uncertainty compared to normal examples under $\mathcal{A}'$. As observed in our empirical study in the following sections, this trade-off is actually helpful for detecting out-of-distribution faults. In other words, we have made the model more sensitive in detecting deviations from the distribution of normal data by using an alternative learning algorithm $\mathcal{A}'$ that can suppress the uncertainty of the normal data on the auxiliary task.

The above information-theoretical analysis motivates us to incorporate an auxiliary learning task into the existing classification model to get improved sensitivity to potential out-of-distribution faults. In the upcoming sections, we will describe how we design an augmented neural network architecture by incorporating reconstruction (autoencoding) as an auxiliary learning task to achieve the desired trade-off.

\begin{figure}[tb]
  \centering
  \includegraphics[width=0.5\linewidth]{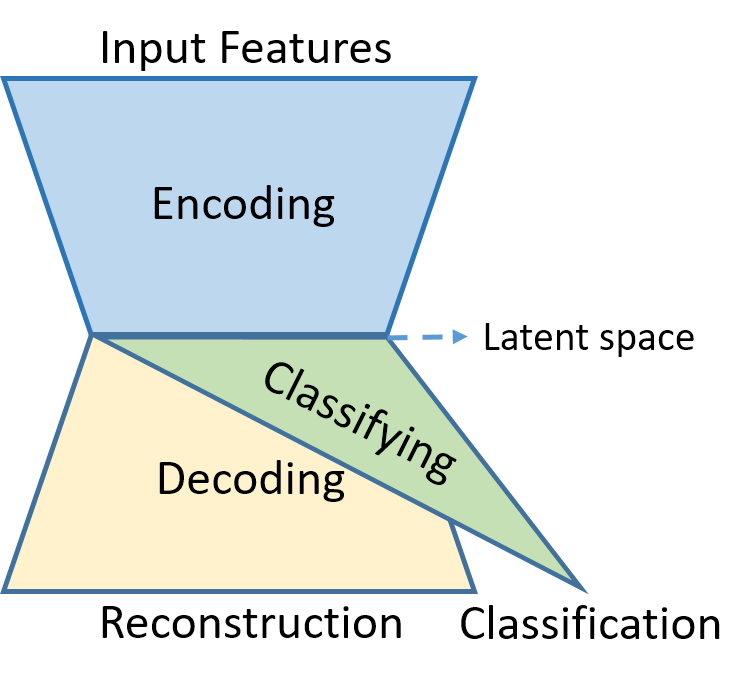}
  \caption{Structure of the propose augmented model.}
  \label{fig:network-structure}
\end{figure}

\section{Methodology}\label{sec:method}

\subsection{Monte Carlo Dropout}

Dropout~\cite{srivastava2014dropout} is a powerful regularization technique to prevent over-fitting neural network parameters.
In effect, the dropout technique provides an inexpensive approximation to training and evaluating an ensemble of exponentially many neural networks. The dropout mechanism offers a way to incorporate intrinsic randomization into neural network models. Recently, Gal and Ghahramani proposed using MC-dropout~\cite{gal2016uncertainty} to estimate a neural network's prediction uncertainty by using dropout at test time. The uncertainty estimates are obtained by repeatedly ``sampling'' the outputs of a dropout model $M$ given the same input $x$. Suppose we have obtained $T$ i.i.d. sampled outputs $\hat{z}^{(1)}, \hat{z}^{(2)}, \ldots, \hat{z}^{(T)}$ (output probabilities from the softmax layer). Their \textit{predictive mean} $\mu=\frac{1}{T}\sum_{k=1}^T\hat{z}^{(k)}$ can be understood as the expected output given input $x$, and the \textit{predictive variance} $\sigma=\frac{1}{T}\sum_{k=1}^T\left(\hat{z}^{(k)}-\mu\right)^2$ can be used to measure the \textit{confidence} of $M$ in its prediction.

\subsection{Augmented Classification Network}
As motivated in the previous section, we augment a regular classification network by adding a decoding pathway to an intermediate layer of the classification network. We illustrate the network structure in Figure~\ref{fig:network-structure}.

The resulting augmented network now has two output pathways, a classifying pathway that aims to output the correct label for an input to the encoding pathway, and an additional decoding pathway whose goal is to reconstruct the original input from the latent space representation. Our model therefore embodies the functionality of both a classifier and an autoencoder. To get satisfactory results from both pathways, we expect the network to learn meaningful representations that are conducive to both tasks at the latent space where the two pathways diverge. To encourage the separation of fault data points from normal data points in the latent space, we use a small classifying pathway and add dropout layers only in the encoding pathway.
A small classifying pathway with limited capacity motivates the network to learn a clear-cut decision boundary between the normal points and the fault points in the latent space. In addition, the random dropout in the encoding pathway also improves the separation between different classes, so that the learned decision boundary will be robust against the stochastic latent space embedding produced by the encoding pathway.


The inclusion of a decoding pathway may benefit the detection of out-of-distribution faults in another aspect. As we know, a classifier trained as a discriminative model tends to find the most discriminative features to distinguish between classes. Much useful information in the original data is thus lost when training a discriminative model; however, such information may be useful for telling out-of-distribution faults from normal data.
Since the decoding pathway encourages more information about the normal class to be preserved in the latent space, these out-of-distribution faults may become more discernible in the latent space, making it possible to detect and even diagnose them.

\paragraph{Loss function} As opposed to an autoencoder or a classifier, during training the augmented network is trained to minimize both the reconstruction loss and the classification loss at the same time. 
We thus use the loss function $\mathcal{L}$ defined as follows
\begin{align}\label{eqn:objective_function}
  \mathcal{L} = \sum^{m}_{i=1}\ell_i
  = \sum^{m}_{i=1}(\ell^\text{clf} + \beta\ell^\text{rec}\cdot\mathbbm{1}_{y_i=0}),\notag
\end{align}
where the loss of each data point $\ell_i$ consists of two parts, the classification loss $\ell^\text{clf}$ and the reconstruction loss $\ell^\text{rec}$. A hyperparameter $\beta$ is introduced to balance the trade-off between the two losses. Since we only want to suppress the prediction uncertainty on the normal data points, we apply the reconstruction loss only on the normal data points. In our experiment to be later described, we used the cross-entropy loss for $\ell^\text{clf}$ and the mean square loss for $\ell^\text{rec}$.

\paragraph{Evaluation}
Since our proposed model can be used both as an MC dropout autoencoder and an MC dropout classifier, its \ac{FDD} performance can be evaluated in more than one way. These evaluation metrics to be described shortly can also apply to autoencoders and classifiers, and will be used in the case studies for comparing the performance of these networks. In our experiment, we used an MC dropout autoencoder and an MC dropout classifier (hereafter referred to as ``autoencoder'' and ``classifier'' for brevity) as benchmarks to show the performance gain from using our proposed augmented model. For a fair comparison, the two benchmark models share the same components (pathways) as the augmented model.


\subsection{Binary Classification (Fault Detection)}

In fault detection tasks, we need a model to tell whether (or how likely) an example $x_i$ is a fault. Both the decoding pathway and the classifying pathway can be used for this task.

\paragraph{Anomaly scores}
To measure how significant an input example $x_i$ exhibits an anomalous behavior at the output node of a two-class classifying pathway, we define the \textit{anomaly score} of input data $x_i$ as below,
\begin{align}
  s_j^\text{clf}(x_i) =
  \begin{cases}
    \mu_{ij}^\text{clf} + \sigma_{ij}^\text{clf},&j\neq 0,\\
    1 - \mu_{ij}^\text{clf} + \sigma_{ij}^\text{clf},&j=0.
  \end{cases} 
\end{align}
A smaller anomaly score implies less uncertainty in the prediction mean and variance.
Without loss of generality, let us label the $n+1$ output classes (the normal class and $n$ faults) by $j=0,1,\ldots,n$, where class 0 is the normal class. Because the normal state is signified by an output of 1 at the for class 0 output node, the above definition for anomaly scores needs to be specially treated for class 0 output,

For the decoding pathway, the anomaly score of input $x_i$ is defined to be the reconstruction error $e^\text{rec}$ (i.e., the mean-square-error) at the output,
\begin{align}\label{eqn:score_rec}
  s^\text{rec}(x_i) = e^\text{rec}\left(\mu^\text{rec}_i, x_i\right). \notag
\end{align}
Here, $\mu^\text{rec}_i$ is the predictive mean of the reconstruction output produced by the decoding pathway given input $x_i$.

\paragraph{Detection Thresholds}
Ideally, an example belonging to the normal class should give zero anomaly scores for both the classifying pathway and the decoding pathway. In practice, normal examples will still exhibit small anomaly scores, so we need a detection threshold for determining where an input example is faulty or not. To do so, we use a pre-defined $\alpha\in(0,1)$ to determine the detection rate, such that the false positive rate on normal training data is $1-\alpha$. We deem an input example as susceptible if its anomaly score is above the corresponding detection threshold, i.e. $s(x_i)>\Tilde{s}_\alpha$.
The above method of choosing thresholds essentially limits the false positive rate to a given value, thus controlling the costs incurred by false alarms. A similar constant false alarm rate principle~\cite{chen1987detection} has been adopted in adaptive algorithms for radar systems.

\paragraph{Fault detection in multiclass models}

In multiclass models, its fault detection performance is also an important and meaningful metric to evaluate, for two reasons. First, in \ac{FDD} applications, being able to tell the existence fault is by itself meaningful. Second, it is important for an \ac{FDD} model to have the ability to tell a potential deviation from the normal condition. If the input corresponds to an unseen fault type that does not belong to the ones modeled by the classification model, then the detection performance is more valuable than the diagnostic performance.

In the multiclass case, we define a detection threshold in a similar fashion as in the two-class case. Let us use variable $b_{ij}$ to indicate anomalies 
corresponding to the $j^{\text{th}}$ type of fault.
We consider $j$ to be a possible label for input $x_i$ if $b_{ij}=1$. It is possible that an input example $x_i$ gets assigned more than one label, which reflects the classifier's uncertainty about the true label of $x_i$. We define the final predicted label $z(x_i)$ as the disjunction of $b_{ij}$'s,
\begin{align}
  z(x_i) &= \bigvee_{j=0}^{n} b_{ij},~\text{where}~b_{ij}\leftrightarrow s_j^\text{clf}(x_i) > \Tilde{s}_j^\text{clf}.
\end{align}

\subsection{Multiclass Classification (Fault Diagnosis)}
The multiclass case can be seen as a natural extension to the two-class case. One major difference is that the softmax function is used as the activation function in the output layer of the network. A multiclass classification model has the capability to tell what type of fault an input corresponds to (fault diagnosis). We introduce the notion of \textit{diagnostic accuracy} for evaluating how accurately a multiclass model can pinpoint the underlying fault type.

\paragraph{Diagnostic Accuracy}
In a multiclass setting, fault diagnosis is a more difficult task than just detecting the existence of faults.
Let $Y_i=\{j\,\vert\,b_{ij}=1\}$ be the set of predicted labels of the classifier on input $x_i$, and $y_i$ be the ground-truth label of input $x_i$. Note that in our context each example only has one label.  
We define the \textit{diagnostic accuracy} $\delta$ as follows, 
\begin{align}
  \delta(x_i) = \frac{\mathbbm{1}_{y_i\in Y_i}}{\sum_{j=1}^{n}b_{ij}}
  \begin{cases}
    =0,&y_i\not\in Y_i\\
    =1,&y_i\in Y_i~\text{and}~\vert{Y_i}\vert=1\\
    \in(0,1),&y_i\in Y_i~\text{and}~\vert{Y_i}\vert>1
  \end{cases}
\end{align}
where the denominator $\sum_{j=1}^{n}b_{ij}$ is the total number of detected fault labels, and the numerator $\mathbbm{1}_{y_i\in Y_i}$ indicates whether the true label is correctly detected. The higher the diagnostic accuracy, the more accurately the classification result can pinpoint the true underlying system health status.
It is worthy to note that $j\neq 0$ is excluded in the summation in the denominator. In other words, the diagnostic accuracy will not be discounted if class 0 (the normal class) is included in the set of predicted labels $Y_i$ as long as the correct label is also included. We believe it is an acceptable and desired behavior for an incipient fault as an intermediate state between the normal class and its corresponding fault class to be suspected and labeled as both by an \ac{FDD} algorithm.
\section{Datasets}\label{sec:dataset}

\begin{table*}[]
\caption{Binary classification accuracy and diagnostic accuracy}
\label{tab:detection-rate}
\resizebox{\textwidth}{!}{%
\begin{tabular}{cccccccc}
\hline
\multicolumn{2}{c}{\multirow{3}{*}{}} & \multicolumn{4}{c}{Classification accuracy} & \multicolumn{2}{c}{Diagnostic accuracy} \\
\multicolumn{2}{c}{} & \multirow{2}{*}{\begin{tabular}[c]{@{}c@{}}Combined model\\ (decoding path)\end{tabular}} & \multirow{2}{*}{\begin{tabular}[c]{@{}c@{}}MC dropout\\ autoencoder\end{tabular}} & \multirow{2}{*}{\begin{tabular}[c]{@{}c@{}}Combined model\\ (classification path)\end{tabular}} & \multirow{2}{*}{\begin{tabular}[c]{@{}c@{}}MC dropout\\ classifier\end{tabular}} & \multirow{2}{*}{\begin{tabular}[c]{@{}c@{}}Combined\\ model\end{tabular}} & \multirow{2}{*}{\begin{tabular}[c]{@{}c@{}}MC dropout\\ classifier\end{tabular}} \\
\multicolumn{2}{c}{} &  &  &  &  &  &  \\ \hline
Thyroid & Normal & \textbf{0.900} & \textbf{0.900} & \textbf{0.900} & \textbf{0.882} & - & - \\
 & Subnormal & \textbf{0.208} & 0.063 & \textbf{0.627} & 0.377 & - & - \\
 & Diseased & 0.910 & 0.860 & \textbf{1.000} & \textbf{1.000} & - & - \\ \hline
Chiller & Normal & 0.937 & 0.944 & 0.942 & 0.936 & - & - \\
 & SL1 & \textbf{0.936} & 0.290 & \textbf{0.618} & 0.503 & \textbf{0.244} & 0.133 \\
 & SL2 & \textbf{0.885} & 0.565 & \textbf{0.716} & 0.703 & \textbf{0.203} & 0.169 \\
 & SL3 & 0.815 & 0.853 & 0.921 & 0.896 & \textbf{0.270} & 0.233 \\
 & SL4 & \textbf{0.796} & 0.319 & 1.000 & 1.000 & 0.280 & 0.311 \\
 & unknown & 0.936 & 0.043 & \textbf{0.999} & 0.133 & - & - \\ \hline
Digits & Zero & 0.947 & 0.947 & \textbf{0.949} & 0.946 & - & - \\
 & Non-zero & \textbf{0.997} & 0.977 & 1.000 & 1.000 & \textbf{0.717} & 0.587 \\
 & Ambiguous & \textbf{0.371} & 0.268 & \textbf{0.530} & 0.492 & \textbf{0.300} & 0.231 \\
 & Out of domain & \textbf{0.999} & 0.997 & \textbf{0.999} & 0.998 & - & - \\ \hline
\end{tabular}%
}
\end{table*}

We selected three datasets from different domains to benchmark the performance of our proposed model. The common trait shared between the three datasets is that they all have some notion of ``incipient faults'' or ``unknown faults''; these out-of-distribution faults will not be represented in the training data. In the case of the hypothyroidism dataset, the resulting model is a binary classification model. Uncertainty information given by the deep learning model will be used to indicate a potential third class--the subnormal condition, which can be seen as an ``incipient fault''. On the chiller dataset, we will train a multiclass classification model for predicting different types of faults. We also tested our approach on the MNIST~\cite{lecun1998mnist} dataset to show that our proposed approach can also work with image data. More details about these datasets will be given below.

\subsection{Thyroid Disease Data}

\looseness -1 We used the ``ANN-thyroid'' dataset from the UCI machine learning repository~\cite{Dua:2019}. The dataset contains clinical data from both normal people and those who have been diagnosed hypothyroidism. Each data point has 21 features, among which 15 are binary and the rest are continuous. We used only the continuous features in our experiment.

The data points are classified into three classes. Besides normal and hypothyroidism, there is also a third class that represents subnormal (mild) hypothyroidism~\cite{quinlan1987simplifying}. The dataset is highly unbalanced; the majority of the data points (about $92\%$) correspond to the normal condition. Among the rest, about 67$\%$ are of the subnormal class. The normal data and subnormal data in this model have some overlap (see Figure~\ref{fig:raw-data-LDA} in the supplemental material), which will cause some difficulties in differentiating the normal data from the subnormal data in classification.

\subsection{RP-1043 Chiller Data}

We used the ASHRAE~RP-1043 Dataset~\cite{comstock1999development} to test out the proposed approach in a multi-class setting. In RP-1043, sensor measurements of a typical cooling system---a 90-ton centrifugal water-cooled chiller---were recorded under both fault-free and various fault conditions. Besides the normal state (NM), We used seven different types of process faults (referred to as FWC, FWE, RL, RO, CF, NC \& EO) from the RP-1043 dataset in our study; see~\cite{jin2019detecting} for a detailed description. Each fault was introduced at four levels of severity (SL1\,-\,SL4, from slightest to severest), except for the EO fault that only has three severity levels. We used the same sixteen features as previous work~\cite{jin2019detecting} did.

Only the normal (SL0) data and the SL4 fault data were used for training the classification models. The less severe SL1 \& SL2 \& SL3 faults were held out as ambiguous examples for testing purpose. We also held out the EO fault to see how the networks will respond to unknown fault examples.

\subsection{MNIST Digits Data}

In this case study, we considered digit-0 images from the MNIST dataset~\cite{lecun1998mnist} to be the ``normal'' class, and four other digits (5, 6, 8 \& 9), which are similar to digit-0 as ``faults''. As with the chiller dataset, we used two types of out-of-distribution examples in our study. The first type is ambiguous digits that resemble two or more digits. For example, there are some digit-0's that are easily mistaken as digit-6's. To generate ambiguous examples that, we used a \ac{VAE} to interpolate between digit 0 and the fault digits. The interpolation was done in the latent space. The other five digits (1, 2, 3, 4 \& 7) were used as ``unknown faults'' in this study.


\section{Experimental Evaluation}\label{sec:experiment}


To examine the performance of our proposed augmented model, we used an autoencoder model and a classifier as benchmark models, and compared their \ac{FDD} performance on both in-distribution and out-of-distribution test data. In this section, we will demonstrate the results on the three datasets described in the previous section.

These three models all share the same encoding pathway, but are different in their output pathways. Both the autoencoder and the classifier can be seen as part of the augmented model: the autoencoder only has a decoding pathway, and the classifier only has a classification pathway. The augmented model has both output pathways that are of the same structures as those in the classifier and the autoencoder. In our model, we only add dropout layer to the encoder pathway. The neural networks used in our experimental study were implemented in Keras~\cite{chollet2015keras}. We set $\beta=1$ for in all of our experiments.


We evaluate the three models' \ac{FDD} performances in terms of their binary classification accuracy and diagnostic accuracy.
The autoencoder model is evaluated by its fault detection performance in terms of binary classification accuracy. The classification model is evaluated by its diagnostic accuracy, in addition to binary classification accuracy. Because the augmented model has two output pathways, it can be compared with the above two models separately depending on which output pathway we focus on. 


\subsection{Thyroid Dataset}

We built a network with \ac{FC} layers for the thyroid dateset, whose latent space had two dimensions. The network had three \ac{FC} layers in its encoding pathway and decoding pathway and two \ac{FC} layers in its classifying pathway. We added a dropout layer after each \ac{FC} layer in the encoding pathway. Among three categories of data, we chose the the normal and the diseased data and randomly divided them into the training set and the test set. The whole training set was used to train the augmented model and the classification model while the autoencoder was trained only with the normal data. For the augmented model, reconstruction was presumably a more difficult task than classification. We found it easier to reconcile the two objectives if we pre-trained the model only with the reconstruction loss as a warm-up. In our experiment, the augmented model was pre-trained for 20 epochs and then trained with the joint objective for another 100 epochs.
 
\begin{figure}[tb]
\centering
    \begin{subfigure}[t]{0.49\linewidth}
    \centering
    \includegraphics[width=0.99\linewidth]{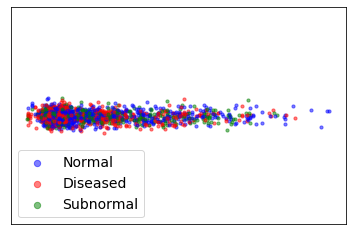}
    \caption{Thyroid: autoencoder}
    \label{fig:thyroid-latent-autoencoder}
    \end{subfigure}
    \begin{subfigure}[t]{0.49\linewidth}
    \centering
    \includegraphics[width=0.99\linewidth]{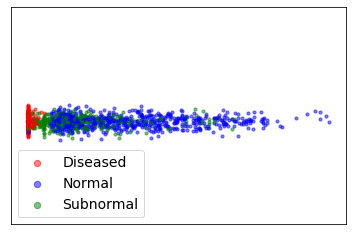}
    \caption{Thyroid: augmented model}
    \label{fig:thyroid-latent-combined}
    \end{subfigure}
    
    \begin{subfigure}[t]{0.49\linewidth}
    \centering
    \includegraphics[width=0.99\linewidth]{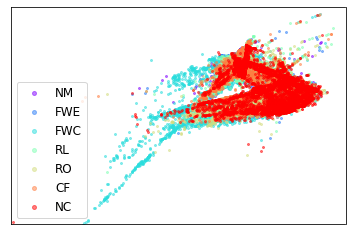}
    \caption{Chiller: autoencoder}
    \label{fig:chiller-latent-autoencoder}
    \end{subfigure}
    \begin{subfigure}[t]{0.49\linewidth}
    \centering
    \includegraphics[width=0.99\linewidth]{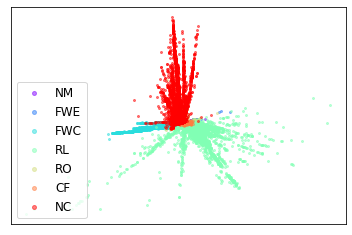}
    \caption{Chiller: augmented model}
    \label{fig:chiller-latent-combined}
    \end{subfigure}
    
    \begin{subfigure}[t]{0.49\linewidth}
    \centering
    \includegraphics[width=0.99\linewidth]{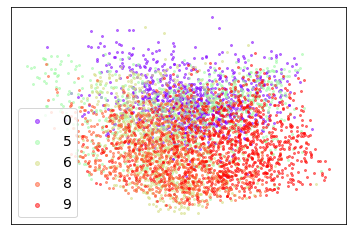}
    \caption{Digits: autoencoder}
    \label{fig:MNIST-latent-autoencoder}
    \end{subfigure}
    \begin{subfigure}[t]{0.49\linewidth}
    \centering
    \includegraphics[width=0.99\linewidth]{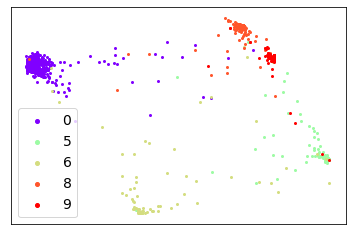}
    \caption{Digits: augmented model}
    \label{fig:MNIST-latent-combined}
    \end{subfigure}
\caption{LDA visualizations of the latent space on the three datasets given by the autoencoder and the augmented model.}
\label{fig:latent}
\end{figure}

We calculate the performance metrics of all three models; see Table~\ref{tab:detection-rate} for a comparison. Because of the overlaps between the normal and subnormal data, we give the model more tolerance to false positives and chose $\alpha=0.1$. Table~\ref{tab:detection-threshold} shows that our model has a much lower average detection threshold (0.002), compared with that of the classifier (0.005). Lower detection threshold implies both the means and the standard deviations of normal data prediction are closer to zero. In other words, our augmented model is more sensitive in detecting the outliers. From Table~\ref{tab:detection-rate}, it is clear that our model has a better performance than the autoencoder and the classifier in both fault detection and fault diagnosis. The normal data and diseased data are both accurately classified. On subnormal data, the binary classification accuracy of our augmented model (0.627) is significantly higher than that of the classifier (0.377), which also shows that the subnormal data are more likely to be detected.

\begin{table}[]
\caption{Detection thresholds on different datasets}
\label{tab:detection-threshold}
\resizebox{\linewidth}{!}{%
\begin{tabular}{cccc}
\hline
 & Thyorid & Chiller (average) & Digits (average) \\ \hline
Augmented model (decoding path) & 0.009 & 0.019 & 0.040 \\
MC dropout autoencoder & 0.007 & 0.013 & 0.036 \\
Augmented model (classifying path) & 0.002 & 0.007 & 0.010 \\
MC dropout classifier & 0.005 & 0.018 & 0.019 \\ \hline
\end{tabular}%
}
\end{table}

Considering the decoding pathway, the average detection threshold given by our model (0.009) is slightly higher than that of the MC dropout autoencoder (0.007). We believe the reason to be that our model does both classification and reconstruction tasks while the autoencoder only concentrates on normal data reconstruction. We visualize and compare the latent space given by the autoencoder and our augmented model in Figure~\ref{fig:latent}. In the latent space, the data points mostly reside in a straight line. To better visualize the distributions for each cluster, we add some Gaussian noise to make them spread out. Compared with the MC dropout autoencoder, the data with different labels are more separated in the latent space of our augmented model. The improvement is obvious on the subnormal data. Furthermore, from Table~\ref{tab:detection-rate}, our model also shows a much better performance than the autoencoder in fault detection compared with the autoencoder, with the classification accuracy value increasing from 0.063 to 0.208.

\begin{figure*}[tb]
\centering
    \begin{subfigure}[t]{0.23\linewidth}
    \centering
    \includegraphics[height=3.2cm]{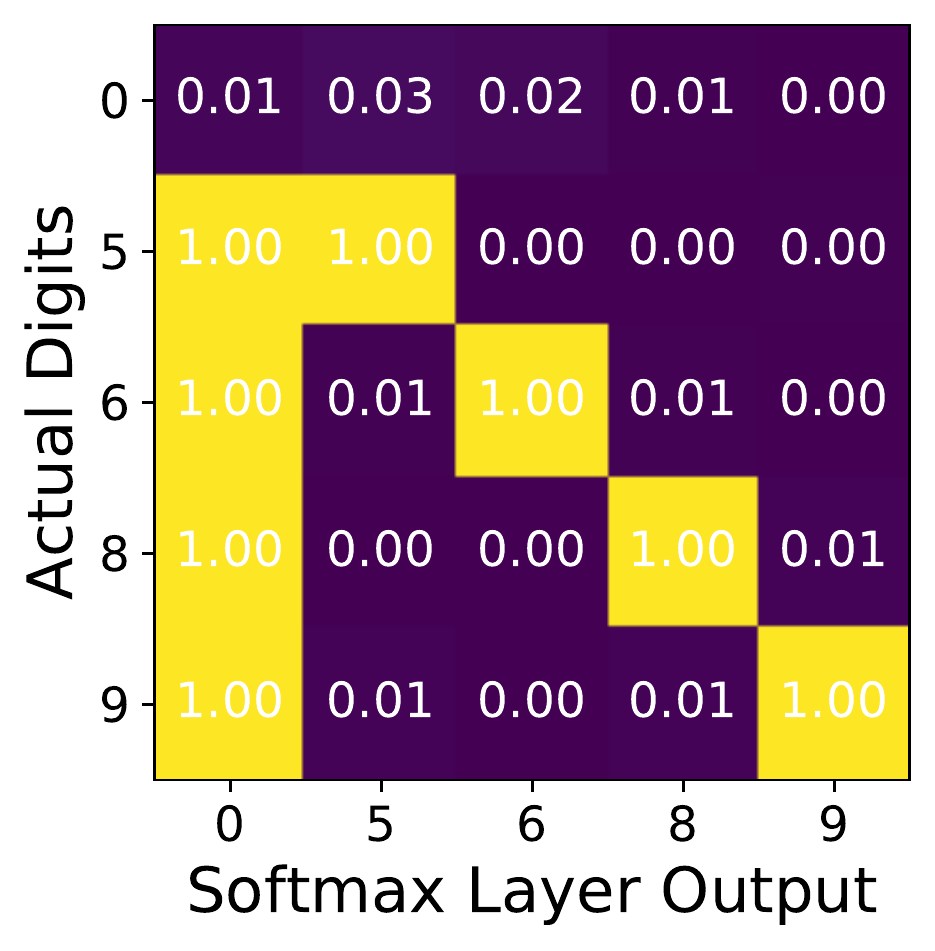}
    \caption{In-distribution faults}
    \label{fig:EMNIST-MC-dropout-classifier}
    \end{subfigure}
    \begin{subfigure}[t]{0.23\linewidth}
    \centering
    \includegraphics[height=3.2cm]{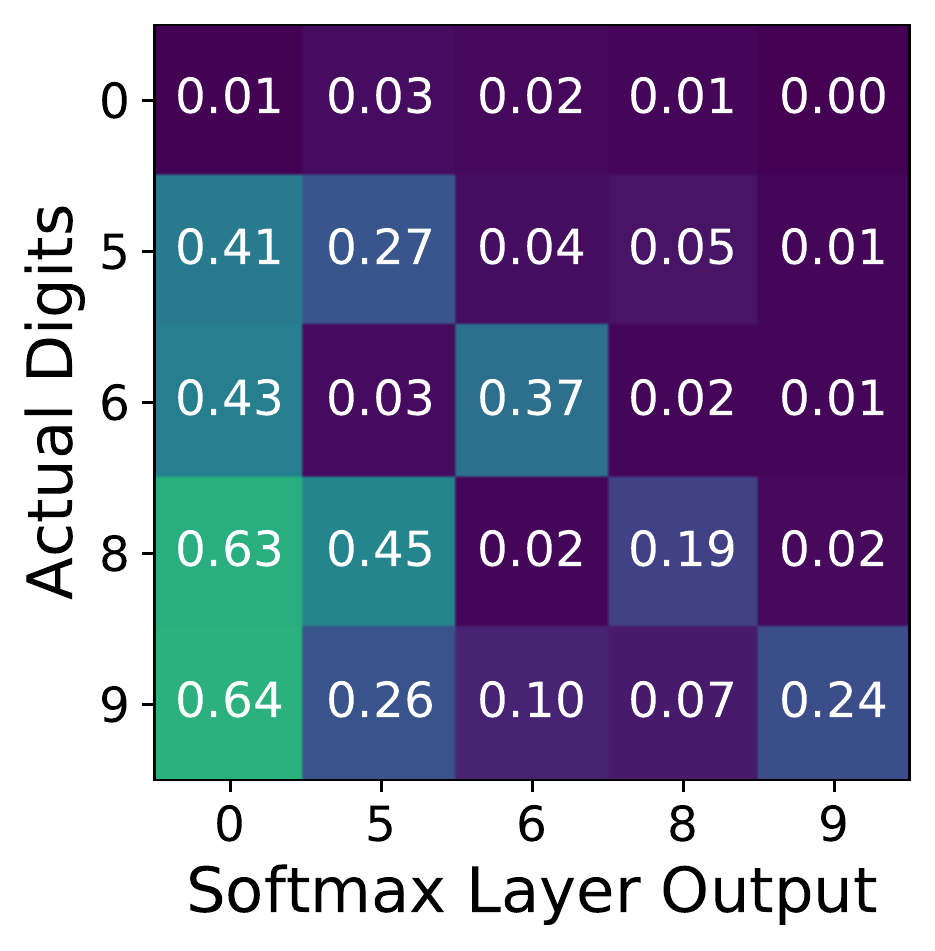}
    \caption{Incipient faults}
    \label{fig:EMNIST-latent-MC-dropout-classifier-incipient-faults}
    \end{subfigure}
    \begin{subfigure}[t]{0.23\linewidth}
    \centering
    \includegraphics[height=3.2cm]{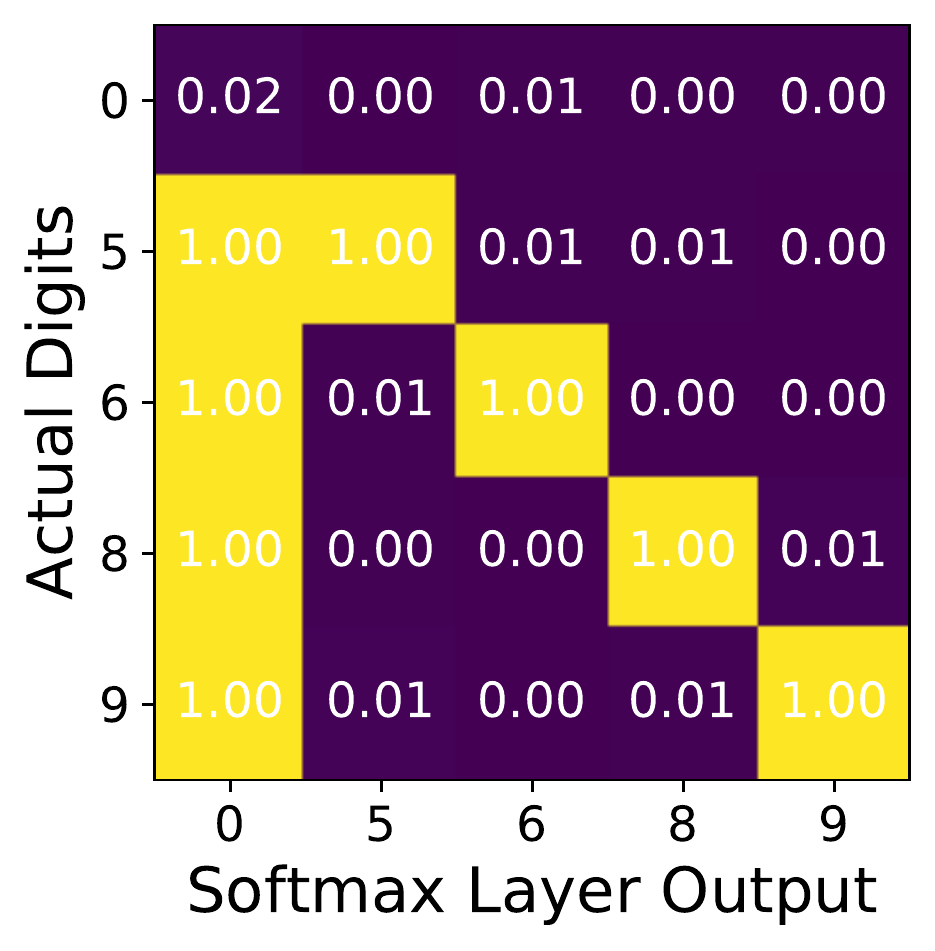}
    \caption{In-distribution faults}
    \label{fig:EMNIST-Combined-model}
    \end{subfigure}
    \begin{subfigure}[t]{0.23\linewidth}
    \centering
    \includegraphics[height=3.2cm]{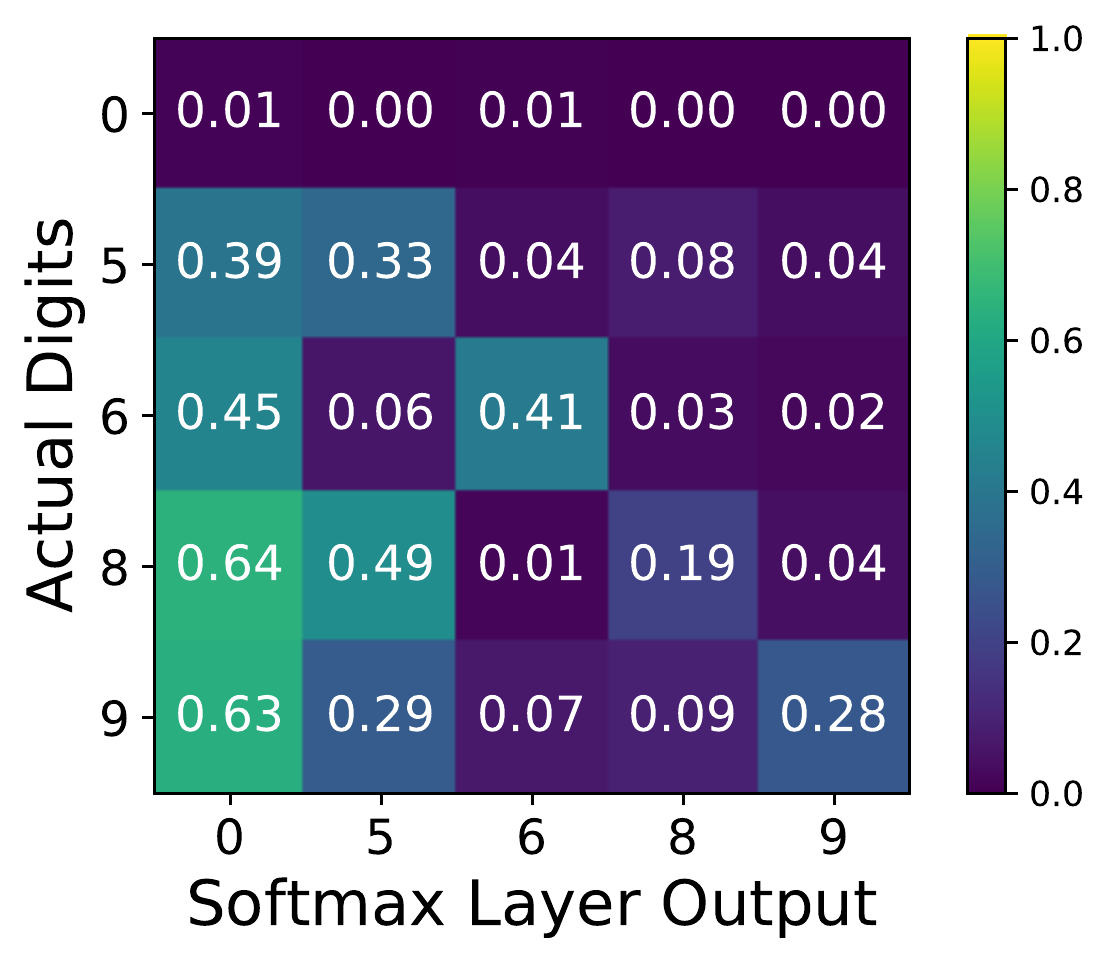}
    \caption{Incipient faults}
    \label{fig:EMNIST-latent-Combined-model-incipient-faults}
    \end{subfigure}
\caption{Anomaly scores on the digits dataset. (a)(b) MC dropout classifier, and (c)(d) augmented model.}
\label{fig:Score-EMNIST}
\end{figure*}

\subsection{Chiller Dataset}

The chiller dataset was also trained and tested by a \ac{FC} neural network, whose latent space had four dimensions. The encoding pathway, decoding pathway and classifying pathway all had three \ac{FC} layers and each \ac{FC} layer in the encoding pathway was followed by a dropout layer. We randomly chose $50\%$ of the data as training set and rest as test set. The normal data and the severe fault (SL4) data were used to train the augmented model and the classifier. For the autoencoder, we trained it only with the normal data. The augmented model was trained for 200 epochs with 40 pre-training epochs and the other two models were both trained until convergence.

\looseness -1 From Table~\ref{tab:detection-rate}, compared with the MC dropout classifier, it is clear that the binary classification accuracy given by the classification pathway in our augmented model is much higher, especially on incipient faults and unknown faults, which means our model performs better in fault detection. The diagnostic accuracy of incipient faults (SL1 to SL3) also sees improvements, which shows that the incipient faults are more likely to be correctly diagnosed. Similar to the thyroid dataset, the detection threshold of our model in chiller dataset (see Table~\ref{tab:detection-threshold}) also decreases compared with the other two models, representing higher sensitivity of our model in detecting potential anomalies.

Binary classification accuracy given by the decoding pathway of our model also has improvements when compared with that given by MC dropout autoencoder. It is noteworthy that the binary classification accuracy has a significant increase in SL1 (0.290 to 0.936), SL2 (0.565 to 0.885) and unknown faults (0.433 to 0.936). Similar to the thyroid experiment, we used a \ac{LDA} visualization on the latent space. As shown in Figure~\ref{fig:latent}, different data clusters in the latent space of our augmented model are more dispersed compared with those of the autoencoder. The incipient faults are partially separated from the normal data, which will benefit our fault detection (see Figure~\ref{fig:latent-space-chiller} in the supplemental material). In addition, the average anomaly scores of our augmented model and the classifier (see Figure~\ref{fig:Ds-chiller} in the supplemental material) can also illustrate the improvements of our model in fault detecting and diagnosis.

\subsection{MNIST Digits Dataset}
We designed a \ac{CNN} to deal with the image data. The encoding pathway had two down-sampling groups, each having two convolution layers and a max-pooling layer. Similarly, the decoding pathway had two up-sampling groups, each having one up-sampling layer and two convolution layers. Between the down-sampling group and the up-sampling group are six \ac{FC} layers with 8 hidden nodes at the bottleneck. The classifying pathway was made up of three \ac{FC} layers. We also added a dropout layer after each \ac{FC} layer in the encoding pathway. 

The augmented model and the MC dropout classifier were trained using both normal data and fault data; for training the autoencoder, only the normal data were used. We trained our model for 200 epochs (40 epochs for pre-training), and trained the other two models until convergence.

We choose $\alpha=0.05$ and compute the performance metrics of the three models, in terms of both binary classification accuracy and diagnostic accuracy. Our model has shown improvements in both accuracy metrics, compared with the MC dropout classifier. As shown in Table~\ref{tab:detection-rate}, the zero and the non-zero data can be almost perfectly classified. 
(please see example prediction results in the supplemental material). Furthermore, our model also has a good performance on the ambiguous data, with the binary classification accuracy reaching 0.530. In multiclass fault diagnosis, our model also shows better diagnostic performance compared with the classifier, with the diagnostic accuracy climbing from 0.587 to 0.717 on non-zero digits, and from 0.231 to 0.300 on ambiguous digits. It can be seen from Table~\ref{tab:detection-threshold} that the average detection threshold of our model is also lower than that of the classifier, which implies that our model is more sensitive than the classifier when detecting outliers.

The average anomaly scores also see improvements of our model in detecting and diagnosing faults. Generally, the anomaly scores calculated from our model are higher than that from the classifier, which represents that our model is more sensitive about the out of distribution faults. The improvements in diagonal value of the matrix means that our model has better performance in diagnosing the faults and incipient faults,

Our augmented model also shows performance improvements in the results from the decoding pathway, compared with that from the autoencoder. As shown in
Table~\ref{tab:detection-rate}, most zero and non-zero digits are both accurately identified. The binary classification accuracy of our model on ambiguous data (0.371) has improved, compared with that of the autoencoder (0.268). Similar to our observations on the previous two datasets, the distributions of data of different classes also become more dispersed in the latent space with our model; see Figure~\ref{fig:latent} for a visualization with \ac{LDA}.


\section{Conclusion}
In this paper, we proposed a novel neural network model for \ac{FDD} applications. In the proposed network structure, an MC dropout classifier is augmented with a decoding pathway; and as a result, the augmented network is trained to perform two tasks simultaneously, classification and reconstruction. We have shown that this combined-objective training can give improved \ac{FDD} performance compared to autoencoders and MC dropout classifiers, especially on out-of-distribution faults. As future work, we plan to conduct a more in-depth theoretical analysis of the proposed method.


\newpage
\clearpage
\bibliography{refs}

\begin{thebibliography}{}

\bibitem[\protect\citeauthoryear{An and Cho}{2015}]{an2015variational}
An, J., and Cho, S.
\newblock 2015.
\newblock Variational autoencoder based anomaly detection using reconstruction
  probability.
\newblock {\em Special Lecture on IE} 2(1).

\bibitem[\protect\citeauthoryear{Chen and Reed}{1987}]{chen1987detection}
Chen, J.~Y., and Reed, I.~S.
\newblock 1987.
\newblock A detection algorithm for optical targets in clutter.
\newblock {\em IEEE Transactions on Aerospace and Electronic Systems}
  (1):46--59.

\bibitem[\protect\citeauthoryear{Chollet and others}{2015}]{chollet2015keras}
Chollet, F., et~al.
\newblock 2015.
\newblock Keras.
\newblock \url{https://keras.io}.

\bibitem[\protect\citeauthoryear{Comstock, Braun, and
  Bernhard}{1999}]{comstock1999development}
Comstock, M.~C.; Braun, J.~E.; and Bernhard, R.
\newblock 1999.
\newblock {\em Development of analysis tools for the evaluation of fault
  detection and diagnostics in chillers}.
\newblock Purdue University.

\bibitem[\protect\citeauthoryear{Dua and Graff}{2017}]{Dua:2019}
Dua, D., and Graff, C.
\newblock 2017.
\newblock {UCI} machine learning repository.

\bibitem[\protect\citeauthoryear{Gal}{2016}]{gal2016uncertainty}
Gal, Y.
\newblock 2016.
\newblock Uncertainty in deep learning.
\newblock {\em University of Cambridge}.

\bibitem[\protect\citeauthoryear{Goodfellow \bgroup et al\mbox.\egroup
  }{2014}]{goodfellow2014generative}
Goodfellow, I.; Pouget-Abadie, J.; Mirza, M.; Xu, B.; Warde-Farley, D.; Ozair,
  S.; Courville, A.; and Bengio, Y.
\newblock 2014.
\newblock Generative adversarial nets.
\newblock In {\em Advances in neural information processing systems},
  2672--2680.

\bibitem[\protect\citeauthoryear{Jin \bgroup et al\mbox.\egroup
  }{2019a}]{jin2019one}
Jin, B.; Chen, Y.; Li, D.; Poolla, K.; and Sangiovanni-Vincentelli, A.
\newblock 2019a.
\newblock A one-class support vector machine calibration method for time series
  change point detection.
\newblock {\em arXiv preprint arXiv:1902.06361}.

\bibitem[\protect\citeauthoryear{Jin \bgroup et al\mbox.\egroup
  }{2019b}]{jin2019detecting}
Jin, B.; Li, D.; Srinivasan, S.; Ng, S.-K.; Poolla, K.; et~al.
\newblock 2019b.
\newblock Detecting and diagnosing incipient building faults using uncertainty
  information from deep neural networks.
\newblock {\em arXiv preprint arXiv:1902.06366}.

\bibitem[\protect\citeauthoryear{Kingma and Welling}{2013}]{kingma2013auto}
Kingma, D.~P., and Welling, M.
\newblock 2013.
\newblock Auto-encoding variational {Bayes}.
\newblock {\em arXiv preprint arXiv:1312.6114}.

\bibitem[\protect\citeauthoryear{LeCun}{1998}]{lecun1998mnist}
LeCun, Y.
\newblock 1998.
\newblock The {MNIST} database of handwritten digits.
\newblock {\em http://yann. lecun. com/exdb/mnist/}.

\bibitem[\protect\citeauthoryear{Leibig \bgroup et al\mbox.\egroup
  }{2017}]{leibig2017leveraging}
Leibig, C.; Allken, V.; Ayhan, M.~S.; Berens, P.; and Wahl, S.
\newblock 2017.
\newblock Leveraging uncertainty information from deep neural networks for
  disease detection.
\newblock {\em Scientific reports} 7(1):17816.

\bibitem[\protect\citeauthoryear{Li \bgroup et al\mbox.\egroup
  }{2019}]{li2019mad}
Li, D.; Chen, D.; Shi, L.; Jin, B.; Goh, J.; and Ng, S.-K.
\newblock 2019.
\newblock {MAD-GAN}: Multivariate anomaly detection for time series data with
  generative adversarial networks.
\newblock {\em arXiv preprint arXiv:1901.04997}.

\bibitem[\protect\citeauthoryear{Quinlan}{1987}]{quinlan1987simplifying}
Quinlan, J.~R.
\newblock 1987.
\newblock Simplifying decision trees.
\newblock {\em International journal of man-machine studies} 27(3):221--234.

\bibitem[\protect\citeauthoryear{Sch{\"o}lkopf \bgroup et al\mbox.\egroup
  }{2001}]{scholkopf2001estimating}
Sch{\"o}lkopf, B.; Platt, J.~C.; Shawe-Taylor, J.; Smola, A.~J.; and
  Williamson, R.~C.
\newblock 2001.
\newblock Estimating the support of a high-dimensional distribution.
\newblock {\em Neural computation} 13(7):1443--1471.

\bibitem[\protect\citeauthoryear{Srivastava \bgroup et al\mbox.\egroup
  }{2014}]{srivastava2014dropout}
Srivastava, N.; Hinton, G.; Krizhevsky, A.; Sutskever, I.; and Salakhutdinov,
  R.
\newblock 2014.
\newblock Dropout: a simple way to prevent neural networks from overfitting.
\newblock {\em The Journal of Machine Learning Research} 15(1):1929--1958.

\bibitem[\protect\citeauthoryear{Tax}{2002}]{tax2002one}
Tax, D. M.~J.
\newblock 2002.
\newblock One-class classification: Concept learning in the absence of
  counter-examples.

\bibitem[\protect\citeauthoryear{Thompson \bgroup et al\mbox.\egroup
  }{2002}]{thompson2002implicit}
Thompson, B.~B.; Marks, R.~J.; Choi, J.~J.; El-Sharkawi, M.~A.; Huang, M.-Y.;
  and Bunje, C.
\newblock 2002.
\newblock Implicit learning in autoencoder novelty assessment.
\newblock In {\em Proceedings of the 2002 International Joint Conference on
  Neural Networks. IJCNN'02 (Cat. No. 02CH37290)}, volume~3,  2878--2883.
\newblock IEEE.

\bibitem[\protect\citeauthoryear{Wang \bgroup et al\mbox.\egroup
  }{2019}]{wang2019self}
Wang, X.; Du, Y.; Lin, S.; Cui, P.; and Yang, Y.
\newblock 2019.
\newblock Self-adversarial variational autoencoder with gaussian anomaly prior
  distribution for anomaly detection.
\newblock {\em arXiv preprint arXiv:1903.00904}.

\bibitem[\protect\citeauthoryear{Zhu and Laptev}{2017}]{zhu2017deep}
Zhu, L., and Laptev, N.
\newblock 2017.
\newblock Deep and confident prediction for time series at {Uber}.
\newblock In {\em 2017 IEEE International Conference on Data Mining Workshops
  (ICDMW)},  103--110.
\newblock IEEE.

\end{thebibliography}
\bibliographystyle{aaai}

\newpage
\clearpage
\section{\textit{Supplemental material}}

Here, we present the additional information as supplemental material about the datasets used in our study and experimental results.

\subsection{Additional Information for the Chiller Dataset}
In the RP-1043 chiller dataset, seven categories of faults are injected into the chiller system, with each fault being introduced at four levels of severity (SL1\,-\,SL4, from the slightest to the severest).
Shown in Table~\ref{tab:fault-characteristics} are the seven categories of faults and their respective normal operation ranges. The condenser fouling (CF) fault was emulated by plugging tubes into the condenser. The reduced condenser water flow rate (FWC) fault and the reduced evaporator water flow rate (FWE) fault were emulated directly by reducing water flow rate in the condenser and evaporator, respectively. The refrigerant overcharge (RO) fault and refrigerant leakage (RL) fault were emulated by increasing and decreasing the amount of refrigerant charge, respectively. The non-condensable in refrigerant (NC) fault was emulated by adding Nitrogen to the refrigerant. The excess oil (EO) fault was emulated by charging more oil than nominal.

\subsection{Additional Visualization for the Thyroid Dataset}
\begin{figure}[b]
  \centering
  \includegraphics[width=0.9\linewidth]{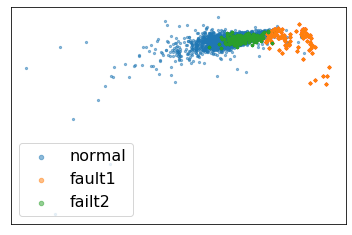}
  \caption{\ac{LDA} visualization of the thyroid dataset}
  \label{fig:raw-data-LDA}
\end{figure}

A visualization of the data points from the thyroid dataset is shown in Fig~\ref{fig:raw-data-LDA}. The plot was created by projecting the subnormal data onto the \acf{LDA} visualization for the in-distribution  data (normal and diseased). The overlap between the normal data and the subnormal data makes it difficult to differentiate between them.

\subsection{Additional Visualization for the Chiller Dataset}
\begin{figure*}[tb]
  \centering
  \begin{subfigure}[t]{0.23\linewidth}
    \centering
    \includegraphics[height=3.5cm]{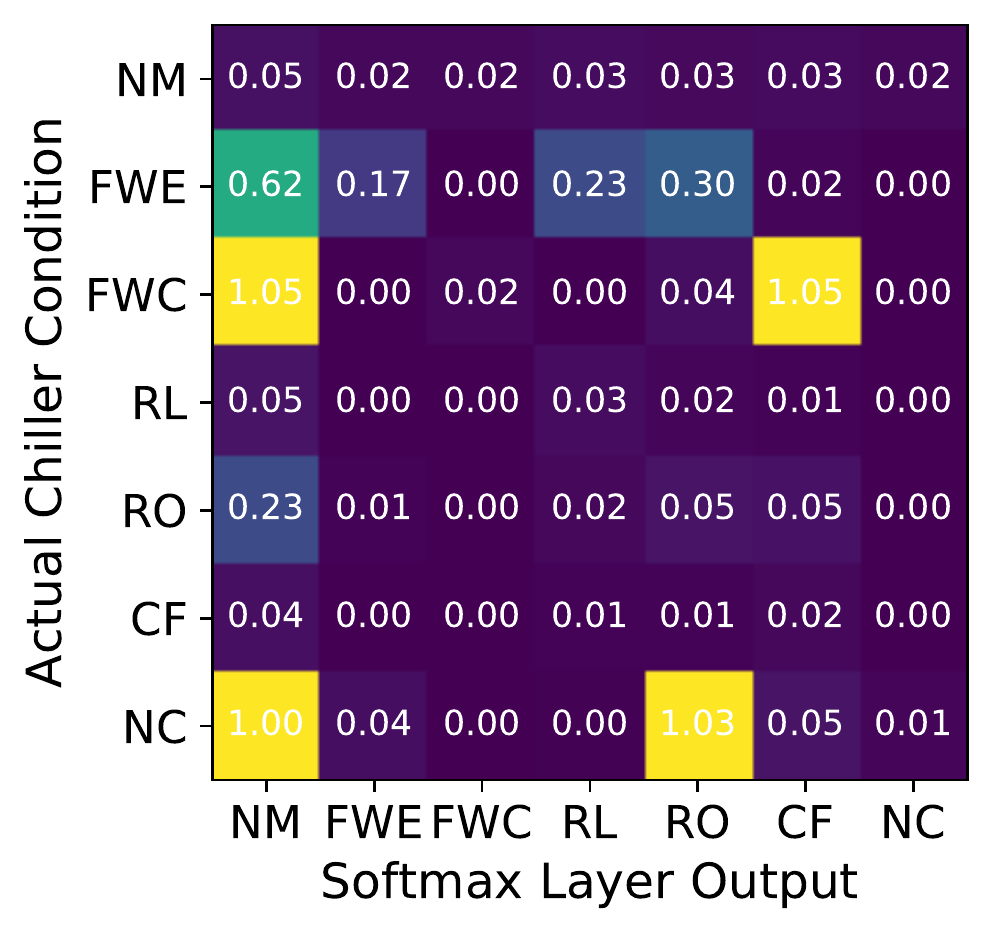}
    \caption{SL1}
    \label{fig:Ds-clf-SL1}
  \end{subfigure}
  \begin{subfigure}[t]{0.23\linewidth}
    \centering
    \includegraphics[height=3.5cm]{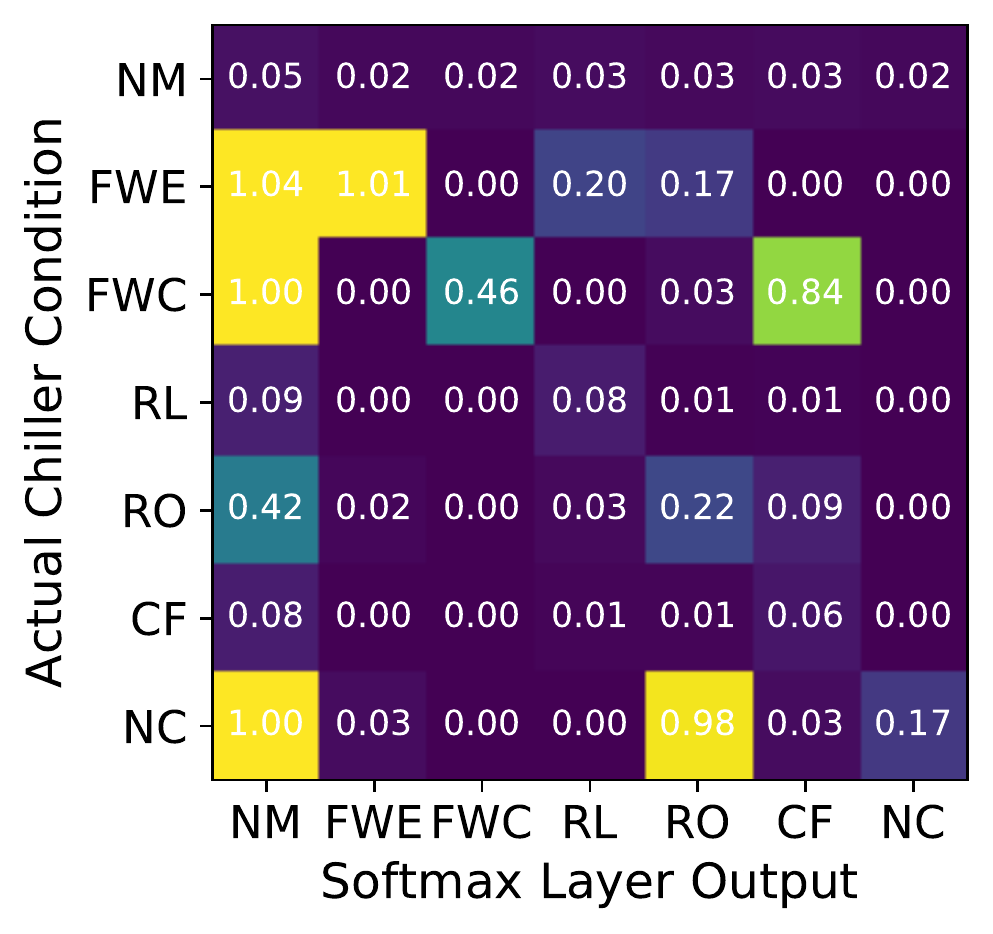}
    \caption{SL2}
    \label{fig:Ds-clf-SL2}
  \end{subfigure}
  \begin{subfigure}[t]{0.23\linewidth}
    \centering
    \includegraphics[height=3.5cm]{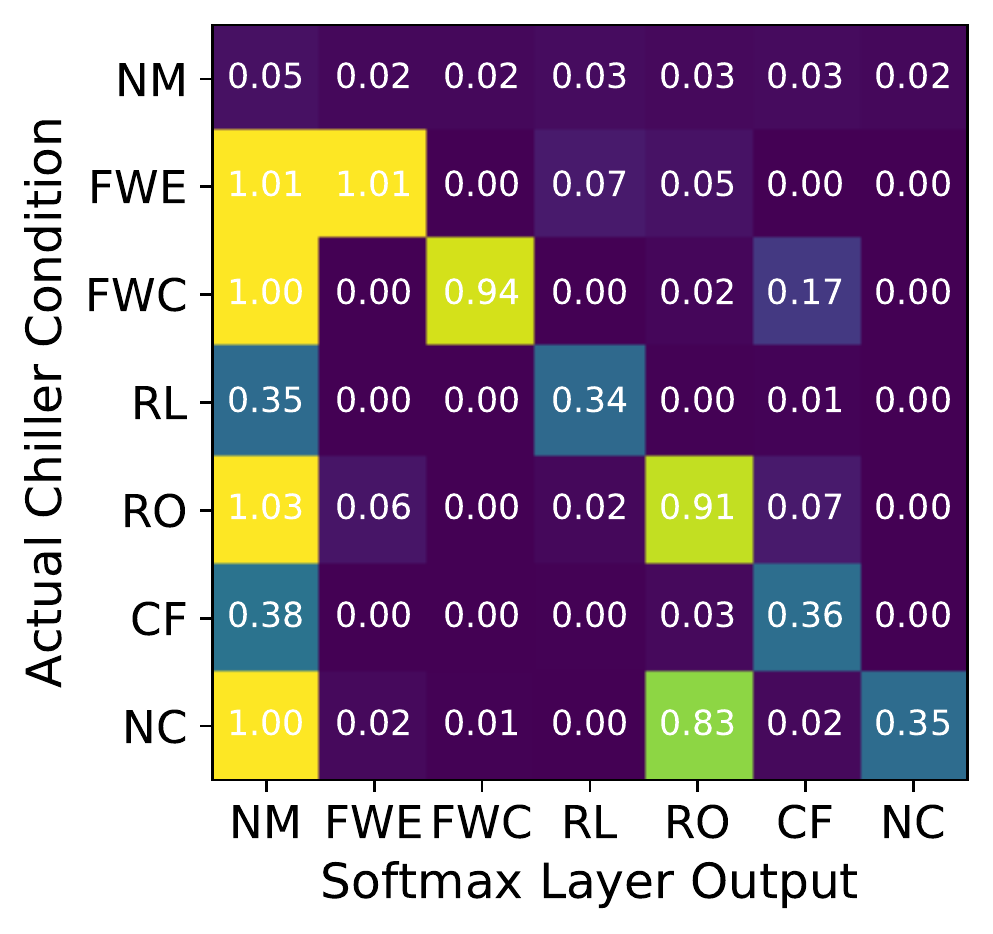}
    \caption{SL3}
    \label{fig:Ds-clf-SL2}
  \end{subfigure}
  \begin{subfigure}[t]{0.23\linewidth}
    \centering
    \includegraphics[height=3.5cm]{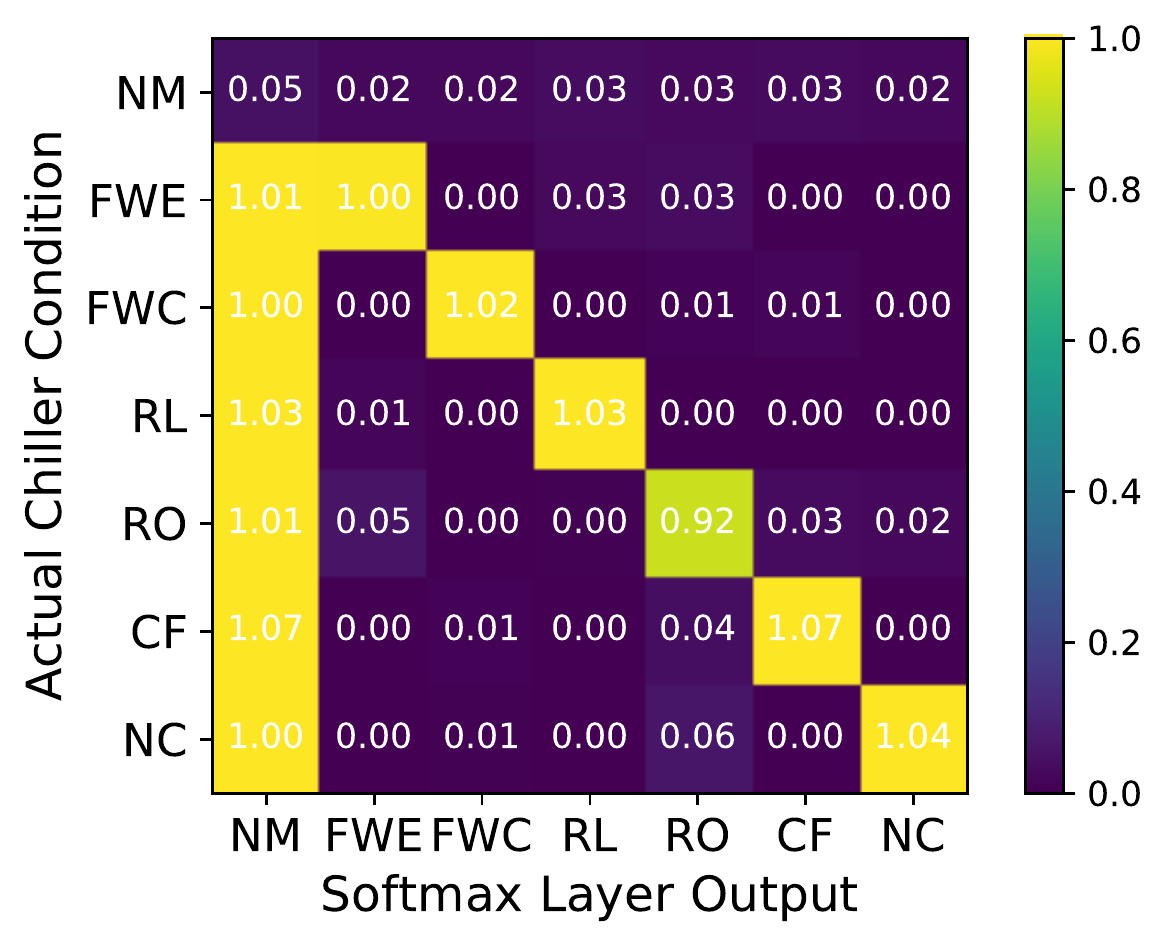}
    \caption{SL4}
    \label{fig:Ds-clf-SL2}
  \end{subfigure}

  \begin{subfigure}[t]{0.23\linewidth}
    \centering
    \includegraphics[height=3.5cm]{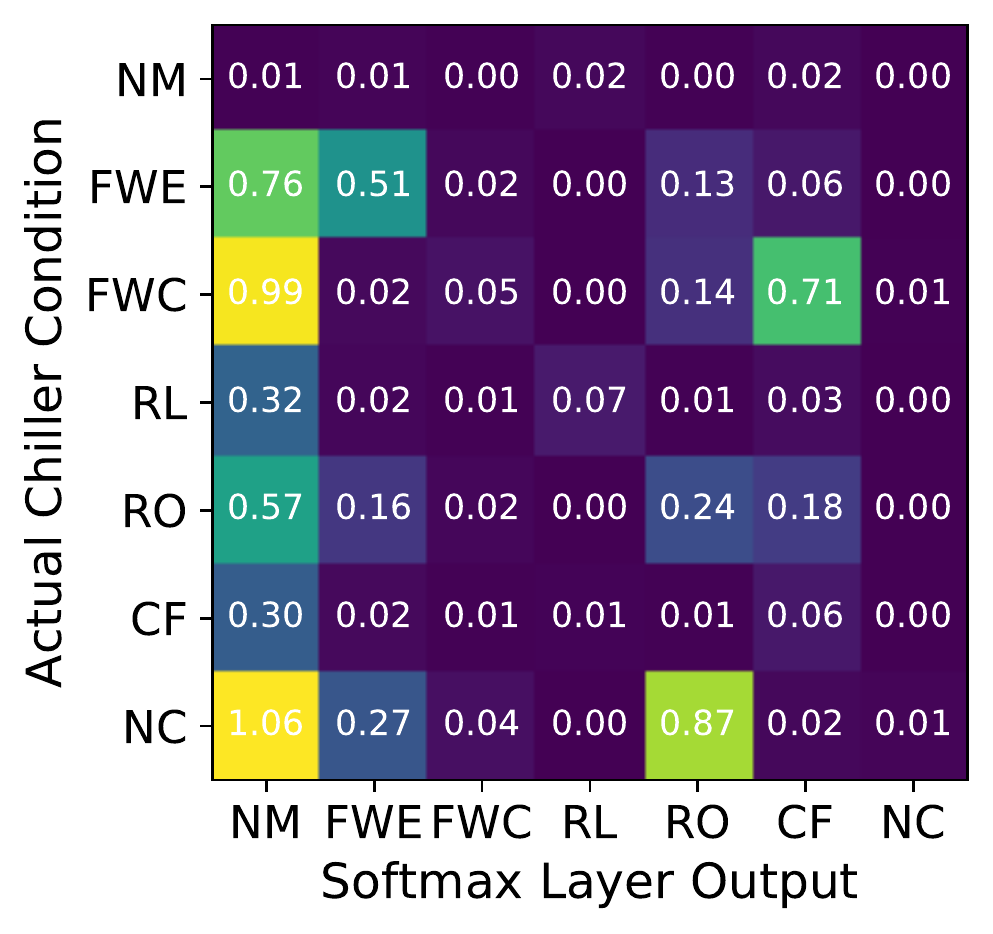}
    \caption{SL1}
    \label{fig:DS-com-SL1}
  \end{subfigure}
  \begin{subfigure}[t]{0.23\linewidth}
    \centering
    \includegraphics[height=3.5cm]{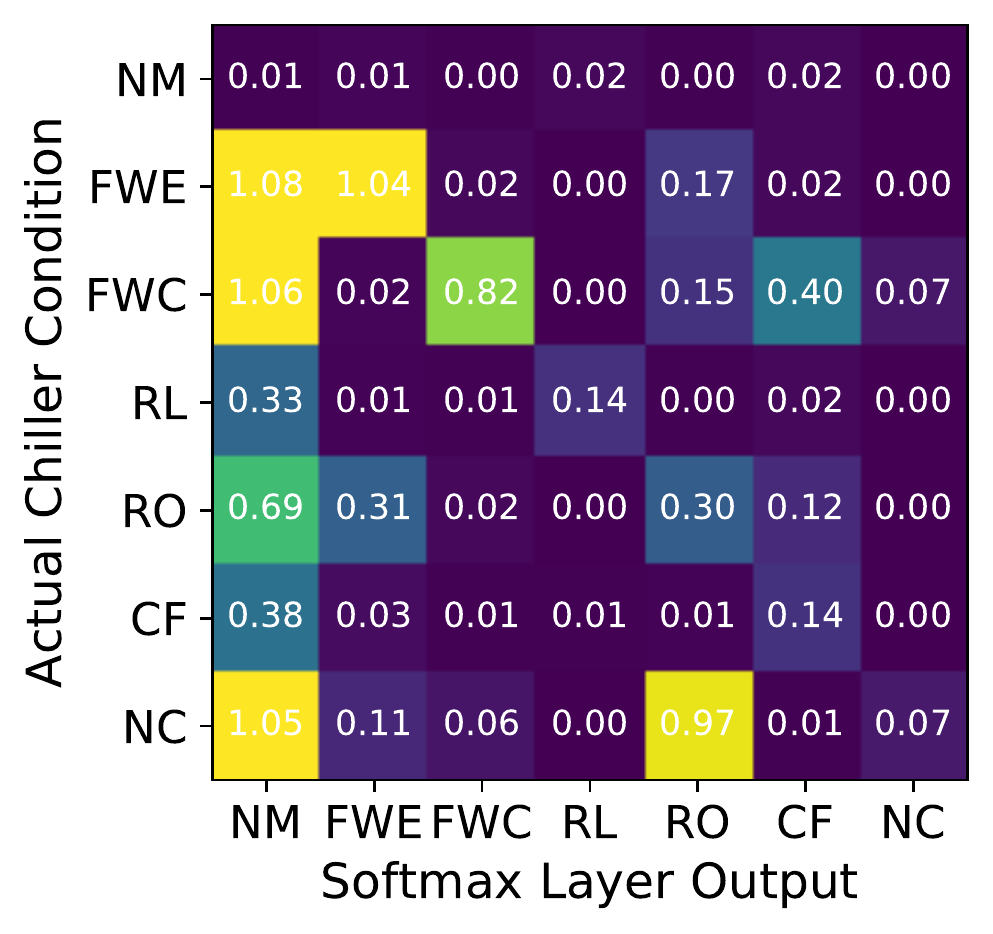}
    \caption{SL2}
    \label{fig:DS-com-SL1}
  \end{subfigure}
  \begin{subfigure}[t]{0.23\linewidth}
    \centering
    \includegraphics[height=3.5cm]{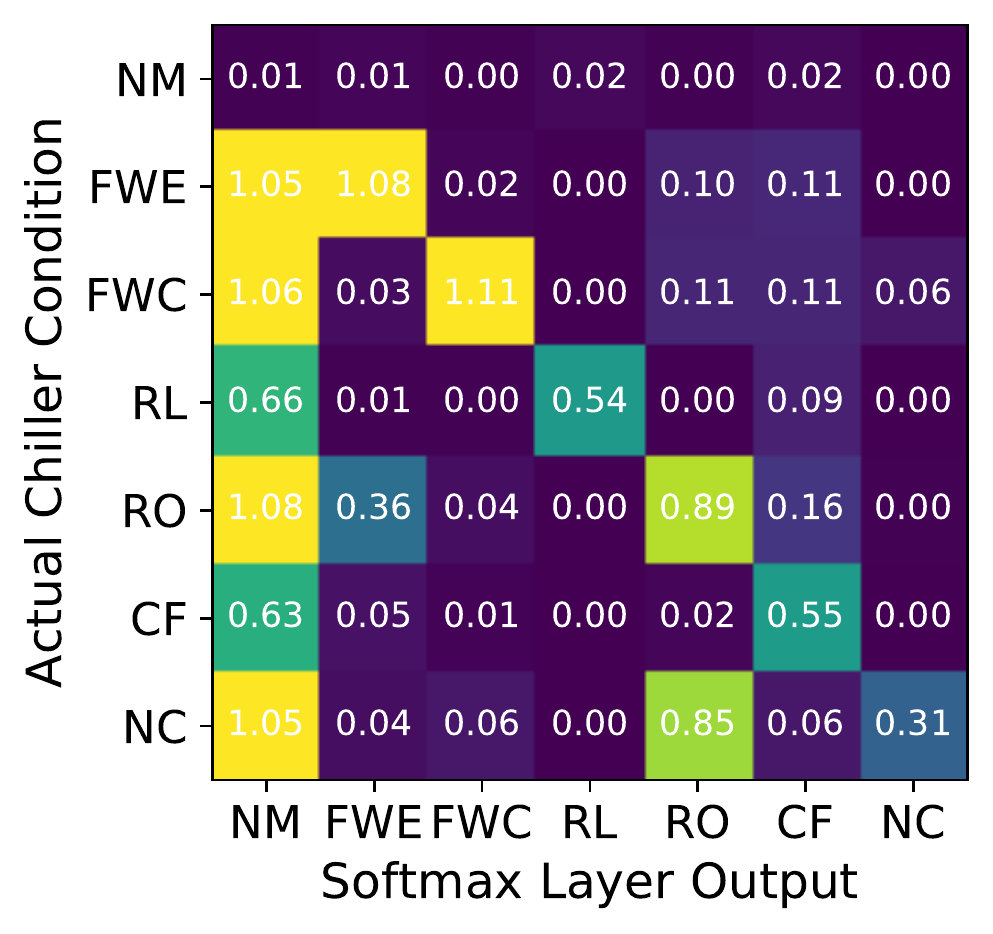}
    \caption{SL3}
    \label{fig:DS-com-SL1}
  \end{subfigure}
  \begin{subfigure}[t]{0.23\linewidth}
    \centering
    \includegraphics[height=3.5cm]{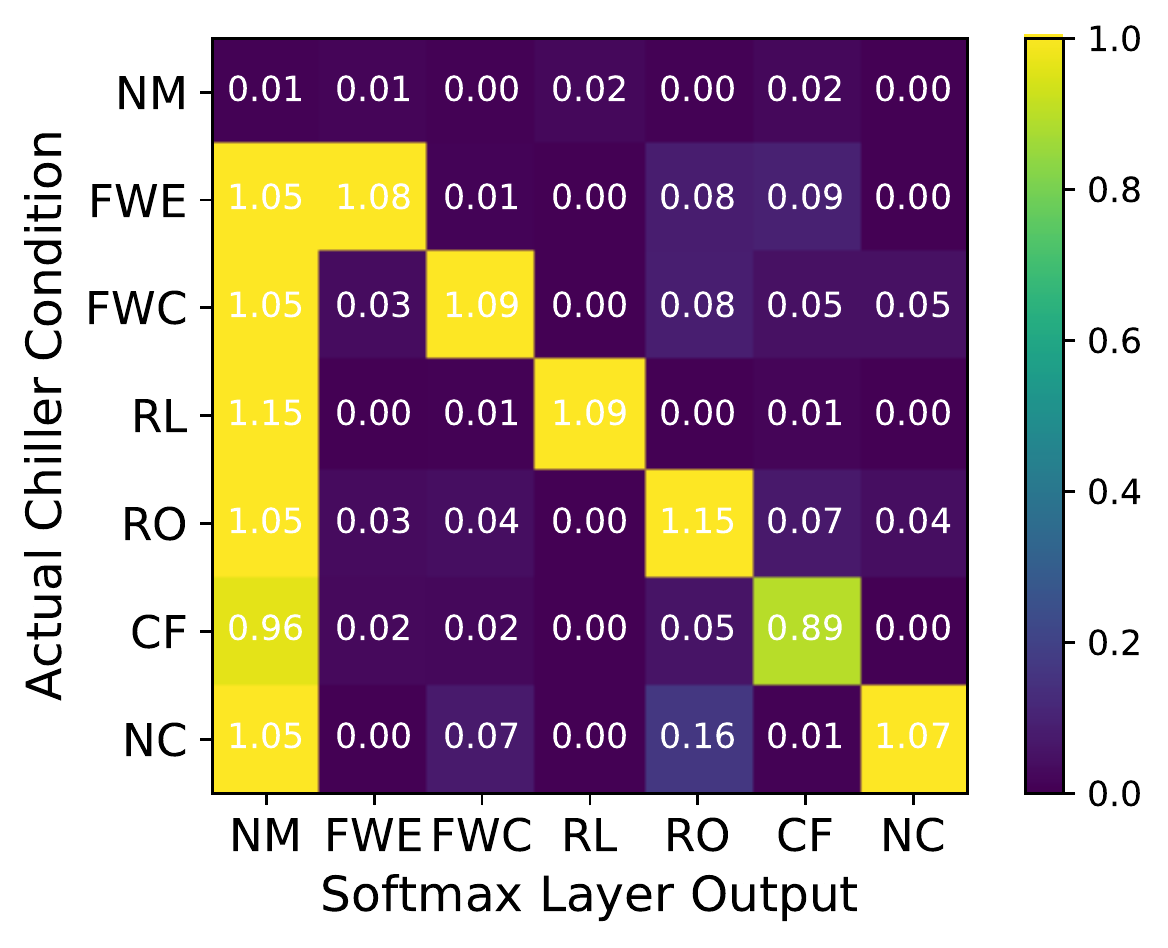}
    \caption{SL4}
    \label{fig:DS-com-SL1}
  \end{subfigure}
  \caption{Anomaly scores on the chiller dataset. (a-d) MC dropout classifier, and (e-h) augmented model.}
  \label{fig:Ds-chiller}
\end{figure*}

Figure~\ref{fig:Ds-chiller} shows the average anomaly scores given by the augmented model and the classifier across different output nodes, and on data of different fault types. This plot is similar to Figure~\ref{fig:Score-EMNIST} in the main text.
Each value in this matrix represents the average anomaly score of one kind of fault at a single output node. It is worth noticing that our model in general gives higher average anomaly scores than the classifier on the diagonal, which shows that our model is not only more accurate but also more sensitive in diagnosing the incipient faults.

\begin{table}[!h]
  \caption{The seven chiller faults used in our study}
  \label{tab:fault-characteristics}
  \centering
  \resizebox{\linewidth}{!}{%
    \begin{tabular}{ll}
      \hline
      \textbf{Fault} & \textbf{Normal Operation} \\ \hline
      Reduced Condenser Water Flow (FWC) & 270 gpm \\
      Reduced Evaporator Water Flow (FWE) & 216 gpm \\
      Refrigerant Leak (RL) & 300 lbs \\
      Refrigerant Overcharge (RO) & 300 lbs \\
      Condenser Fouling (CF) & 164 tubes \\
      Non-Condensables in System (NC) & No nitrogen \\
      Excess Oil (EO) & 22 lbs \\ \hline
    \end{tabular}%
  }
\end{table}

\subsection{Additional Visualization for the Digits Dataset}
\begin{figure*}[tb]
  \centering
  \begin{subfigure}[t]{0.33\linewidth}
    \includegraphics[width=0.95\linewidth]{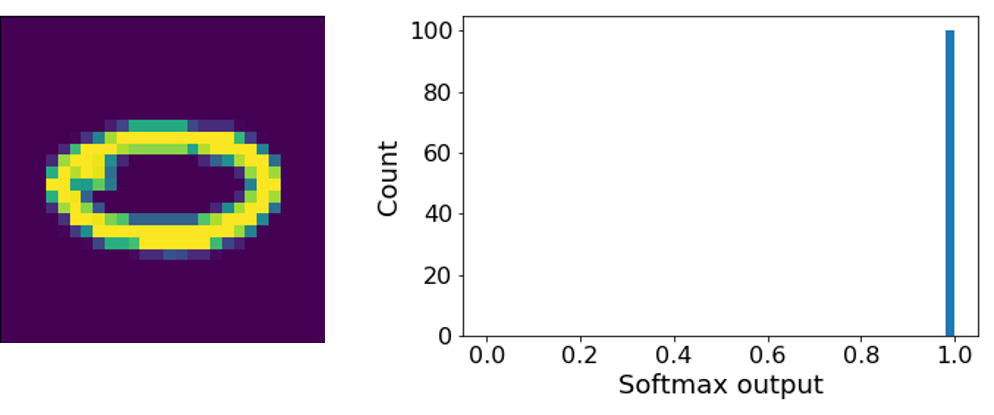}
    \caption{}
    \label{fig:digit-normal}
  \end{subfigure}
  \begin{subfigure}[t]{0.33\linewidth}
    \includegraphics[width=0.95\linewidth]{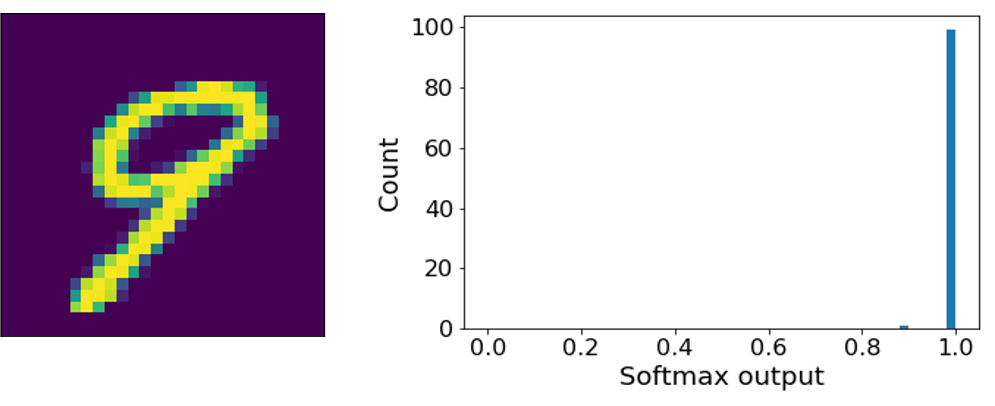}
    \caption{}
    \label{fig:digit-fault-imshow}
  \end{subfigure}
  \begin{subfigure}[t]{0.33\linewidth}
    \includegraphics[width=0.95\linewidth]{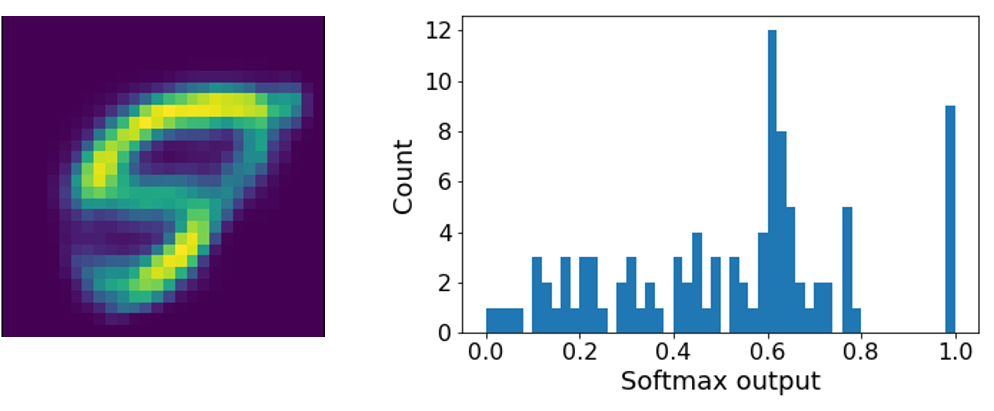}
    \caption{}
    \label{fig:fake-incipient-imshow}
  \end{subfigure}
  \caption{Input image examples (left) and distributions of prediction uncertainty (right) for the MNIST digits dataset. (a) normal data point (digit ``0''), (b) fault data point (digit ``9''), and (c) an ambiguous digit that resembles both 0 and 9.}
  \label{fig:imshow-and-hist}
\end{figure*}

Shown in Figure~\ref{fig:imshow-and-hist} are examples of a normal data point (digit ``0''), a fault data point (digit ``9'') and an ambiguous data point from the MNIST digits dataset, and histograms showing the distributions of their respective prediction outputs from our augmented model under Monte Carlo sampling.
Both digit ``0'' and digit ``9'' are correctly classified into the normal class with small prediction uncertainty, and the prediction uncertainty of digit ``9'' is slightly higher than that of digit ``0''. For the ambiguous digit, much prediction uncertainty can be seen at the output node for class ``9'', showing that the model has difficulty in deciding whether the input image is a ``0'' or ``9''. The results demonstrated by the examples are consistent with our design intent: suppress the uncertainty on normal data and increase the uncertainty on out-of-distribution examples.

\subsection{Visualization for the Chiller Dataset}
The latent space visualization of different severity levels on chiller dataset is shown in Figure~\ref{fig:latent-space-chiller}. It is clear that the incipient fault data and the normal data are becoming less separated in the latent space with the decreasing of the severity level. Some SL1 fault data are highly overlapped with the normal data. Nevertheless, different clusters of data are more dispersed in the latent space using our model than using the MC dropout autoencoder. The normal data are partially separated from the incipient fault data, which offers a good basis for the decoding pathway and classifying pathway to detect and diagnose the incipient faults.

\section{Related Works}

\begin{figure*}[t]
  \centering
  \begin{subfigure}[t]{0.8\textwidth}
    \centering
    \includegraphics[width=0.95\linewidth]{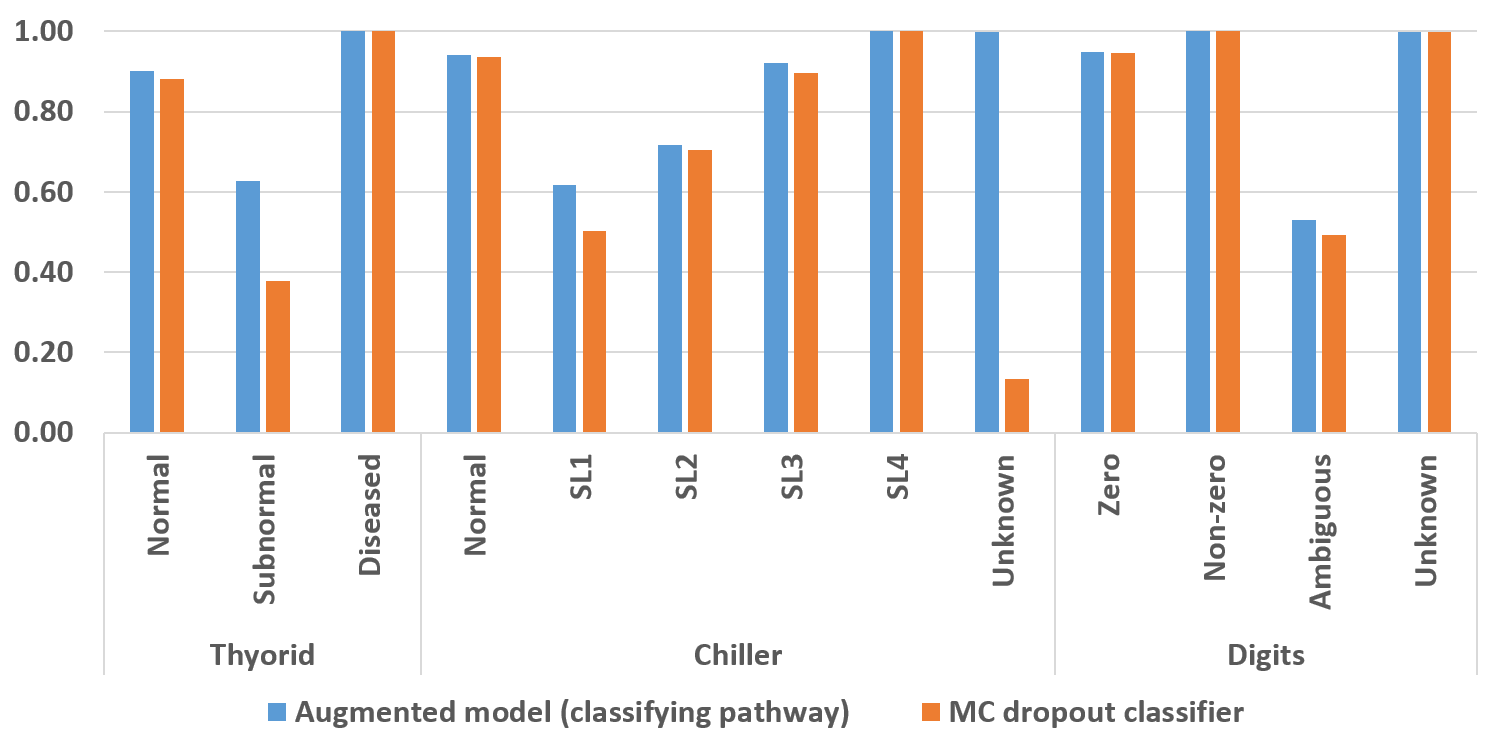}
    \caption{}
    \label{fig:bca-clf&augmented}
  \end{subfigure}

  \begin{subfigure}[t]{0.8\textwidth}
    \centering
    \includegraphics[width=0.95\linewidth]{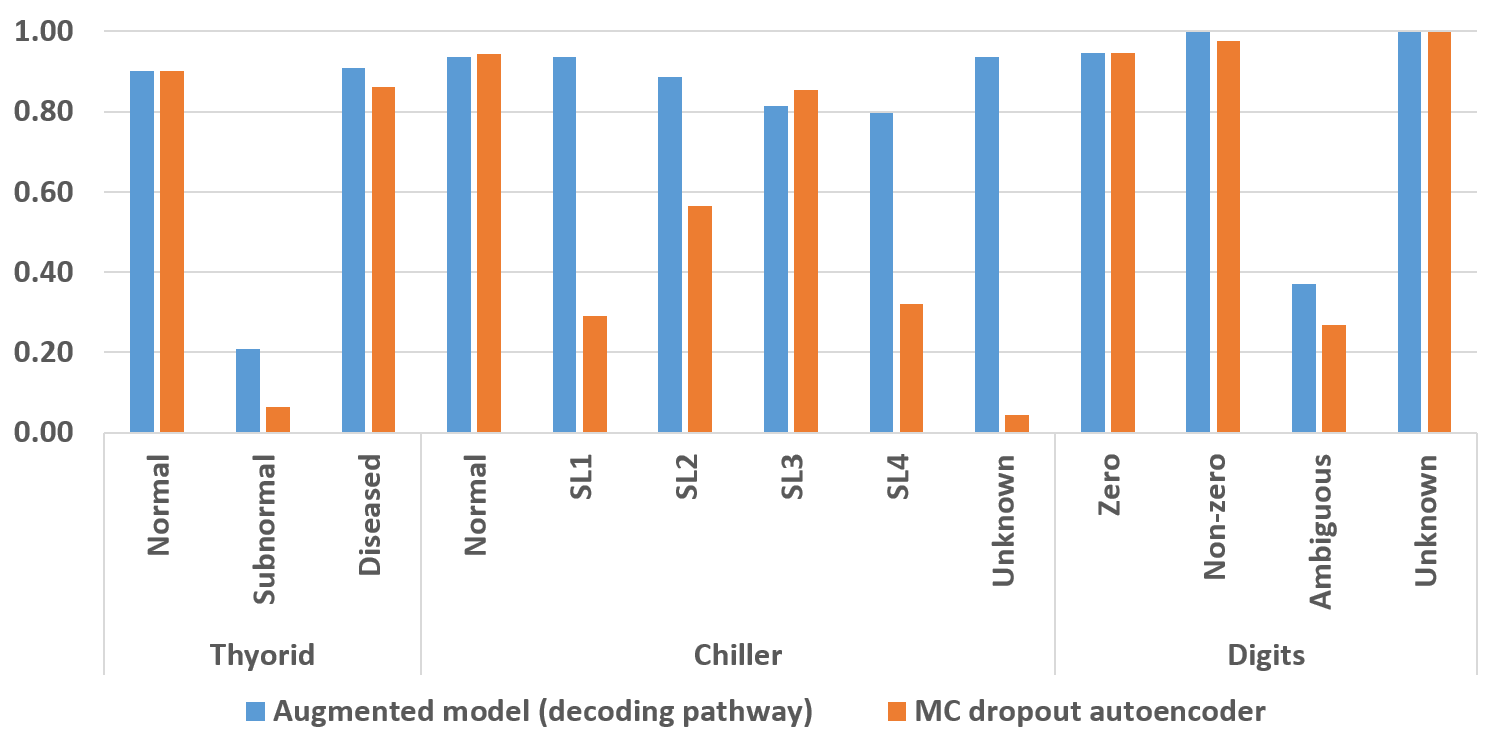}
    \caption{}
    \label{fig:bca-auto&augmented}
  \end{subfigure}
  \caption{The diagram of binary classification accuracy comparison between: (a) classifying pathway in augmented model and the MC dropout classifier (b) decoding pathway in the augmented model and the MC dropout autoencoder}
  \label{fig:latent-space-chiller}
\end{figure*}

Recently, deep generative models such as \ac{VAE}~\cite{kingma2013auto} and \ac{GAN}~\cite{goodfellow2014generative} have become popular in anomaly detection applications.
The main difference between a \ac{VAE} and an autoencoder is that the \ac{VAE} is a stochastic generative model from which a probabilistic measure can be derived for differentiating normal and fault data. In an earlier work~\cite{an2015variational}, the authors used a Monte Carlo method to estimate the reconstruction probability of an input to a \ac{VAE} for identifying faults. The idea of \textit{adversarial training} has also been found useful for anomaly detection, especially in unsupervised settings where adversarial (anomalous) training data are not available. In work~\cite{li2019mad}, a GAN-trained discriminator network learns to detect fake data from real data in an unsupervised fashion. In~\cite{wang2019self}, the authors introduced a self-adversarial training procedure to \ac{VAE}, so that the resulting deep representation not only captures the distribution of normal data but also has discriminative ability against faults.

\begin{figure*}[p]
  \centering
  \begin{subfigure}[t]{0.42\linewidth}
    \centering
    \includegraphics[width=0.95\linewidth]{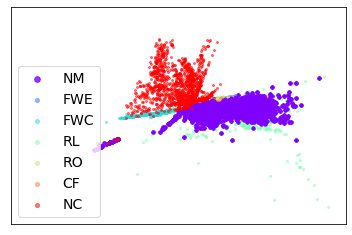}
    \caption{SL1 by the augmented model}
    \label{fig:SL1-combined-latent}
  \end{subfigure}
  \begin{subfigure}[t]{0.42\linewidth}
    \centering
    \includegraphics[width=0.95\linewidth]{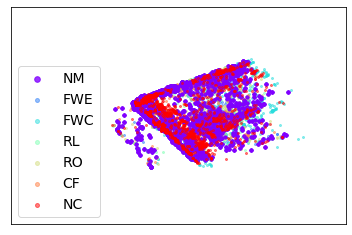}
    \caption{SL1 by the autoencoder}
    \label{fig:SL1-pureae-latent}
  \end{subfigure}

  \begin{subfigure}[t]{0.42\linewidth}
    \centering
    \includegraphics[width=0.95\linewidth]{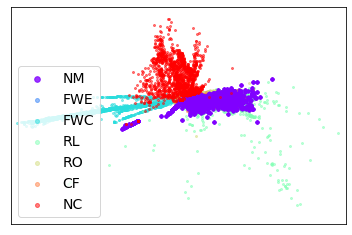}
    \caption{SL2 by the augmented model}
    \label{fig:SL2-combined-latent}
  \end{subfigure}
  \begin{subfigure}[t]{0.42\linewidth}
    \centering
    \includegraphics[width=0.95\linewidth]{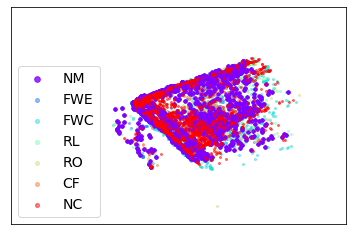}
    \caption{SL2 by the autoencoder}
    \label{fig:SL2-pureae-latent}
  \end{subfigure}

  \begin{subfigure}[t]{0.42\linewidth}
    \centering
    \includegraphics[width=0.95\linewidth]{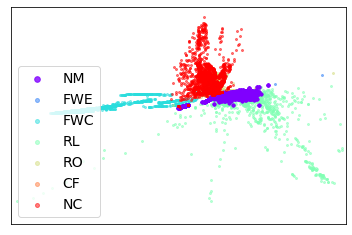}
    \caption{SL3 by the augmented model}
    \label{fig:SL3-combined-latent}
  \end{subfigure}
  \begin{subfigure}[t]{0.42\linewidth}
    \centering
    \includegraphics[width=0.95\linewidth]{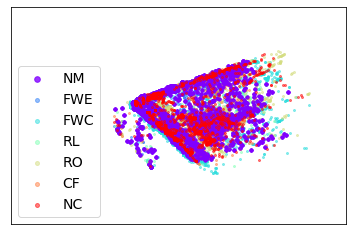}
    \caption{SL3 by the autoencoder}
    \label{fig:SL3-latent}
  \end{subfigure}

  \begin{subfigure}[t]{0.42\linewidth}
    \centering
    \includegraphics[width=0.95\linewidth]{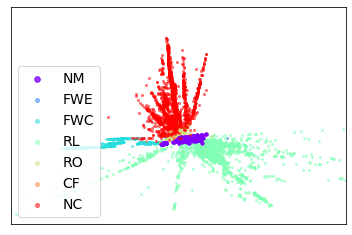}
    \caption{SL4 by the augmented model}
    \label{fig:SL4-combined-latent}
  \end{subfigure}
  \begin{subfigure}[t]{0.42\linewidth}
    \centering
    \includegraphics[width=0.95\linewidth]{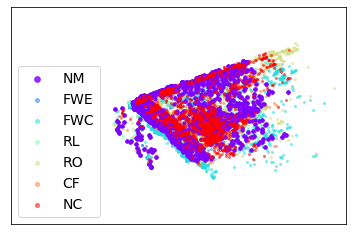}
    \caption{SL4 by the autoencoder}
    \label{fig:SL4-latent}
  \end{subfigure}
  \caption{\ac{LDA} visualization of the the latent space in different severity levels  on the chiller dataset by the augmented model and MC dropout autoencoder.}
  \label{fig:latent-space-chiller}
\end{figure*}


\end{document}